\documentclass{article}

\usepackage{microtype}
\usepackage{graphicx}
\usepackage{subcaption}
\usepackage{booktabs} 
\usepackage[table]{xcolor}
\usepackage{cuted}

\usepackage{hyperref}

\usepackage[preprint]{icml2026}

\usepackage{amsmath}
\usepackage{amssymb}
\usepackage{mathtools}
\usepackage{amsthm}

\usepackage{pifont}

\usepackage{array}
\usepackage{adjustbox}
\newcolumntype{R}[2]{%
    >{\adjustbox{angle=#1,lap=\width-(#2)}\bgroup}%
    l%
    <{\egroup}%
}
\newcommand*\rot{\multicolumn{1}{R{80}{1em}}}

\usepackage[capitalize,noabbrev]{cleveref}

\theoremstyle{plain}

\theoremstyle{definition}

\theoremstyle{remark}

\def\our{EKS}
\def\ourknn{Ray-Traced Gaussian Proximity Search}
\def\ourknnshort{RT-GPS}
\def\ourencoding{Kernel Space Encoding}

\usepackage[textsize=tiny]{todonotes}

\icmltitlerunning{Affine-Equivariant Kernel Space Encoding for NeRF Editing}

\begin{document}

\twocolumn[
  \icmltitle{Affine-Equivariant Kernel Space Encoding for NeRF Editing}



  \icmlsetsymbol{equal}{*}

  \begin{icmlauthorlist}
    \icmlauthor{Miko\l{}aj Zieli\'{n}ski}{pp}
    \icmlauthor{Krzysztof Byrski}{uj}
    \icmlauthor{Tomasz Szczepanik}{uj}
    \icmlauthor{Dominik Belter}{pp}
    \icmlauthor{Przemys\l{}aw Spurek}{uj,ideas}

  \end{icmlauthorlist}
  \icmlaffiliation{pp}{Poznan University of Technology, Institute of Robotics and Machine Intelligence, ul. Piotrowo 3A, Pozna\'{n} 60-965, Poland}
  \icmlaffiliation{uj}{Jagiellonian University, Faculty of Mathematics and Computer Science, \L{}ojasiewicza 6, 30-348, Krakow, Poland}
  \icmlaffiliation{ideas}{IDEAS Research Institute}

  \icmlcorrespondingauthor{Miko\l{}aj Zieli\'{n}ski}{mikolaj.zielinski@put.poznan.pl}

  \icmlkeywords{Machine Learning, ICML}

  \vskip 0.3in
]



\printAffiliationsAndNotice{}  

\begin{strip}
\vspace{-2cm}
     \centering
       \includegraphics[width=\textwidth]{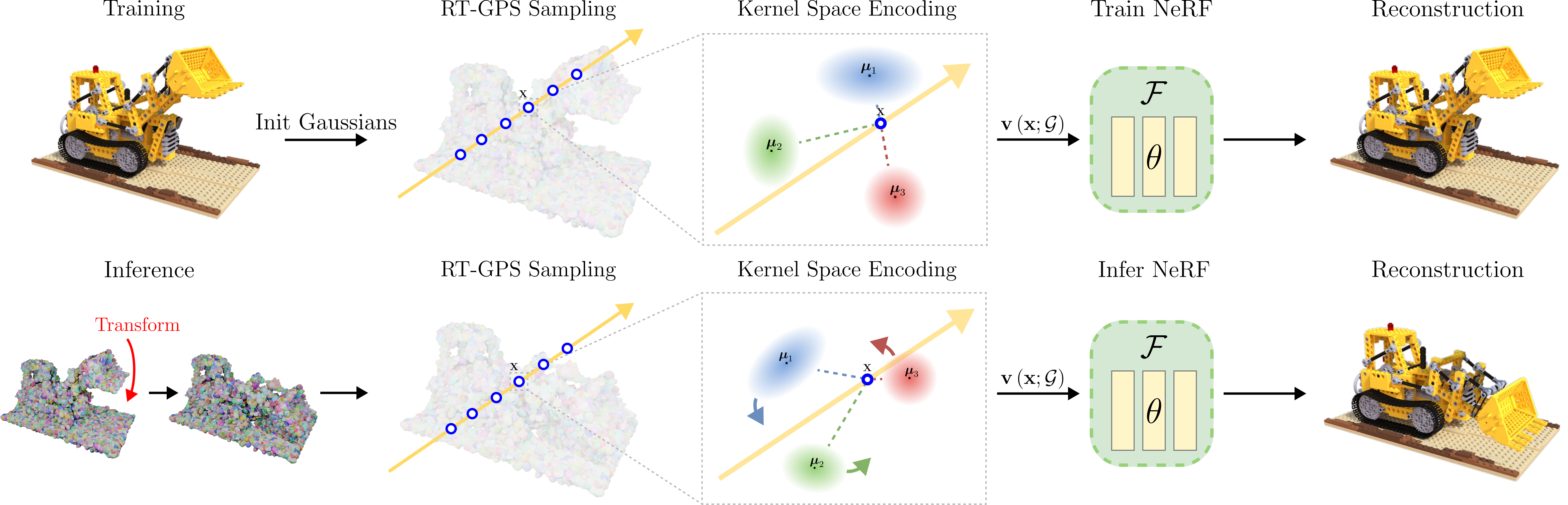} 
     \captionof{figure}{\textbf{\our{} overview.} \our{} represents positional features using spatially localized anisotropic Gaussian kernels, enabling stable and fine-grained interactive editing while maintaining the high-fidelity rendering of Neural Radiance Fields.} 
 \label{fig:teaser} 
 \end{strip}

\begin{abstract}

    Neural scene representations achieve high-fidelity rendering by encoding 3D scenes as continuous functions, but their latent spaces are typically implicit and globally entangled, making localized editing and physically grounded manipulation difficult. While several works introduce explicit control structures or point-based latent representations to improve editability, these approaches often suffer from limited locality, sensitivity to deformations, or visual artifacts. In this paper, we introduce Affine-Equivariant Kernel Space Encoding (\our{}), a spatial encoding for neural radiance fields that provides localized, deformation-aware feature representations. Instead of querying latent features directly at discrete points or grid vertices, our encoding aggregates features through a field of anisotropic Gaussian kernels, each defining a localized region of influence. This kernel-based formulation enables stable feature interpolation under spatial transformations while preserving continuity and high reconstruction quality. To preserve detail without sacrificing editability, we further propose a training-time feature distillation mechanism that transfers information from multi-resolution hash grid encodings into the kernel field, yielding a compact and fully grid-free representation at inference. This enables intuitive, localized scene editing directly via Gaussian kernels without retraining, while maintaining high-quality rendering. The code can be found under (https://github.com/MikolajZielinski/eks)
      
\end{abstract}

\vspace{-0.7cm}

\section{Introduction}

        \begin{figure*}[t]
            \centering
            \includegraphics[width=\textwidth]{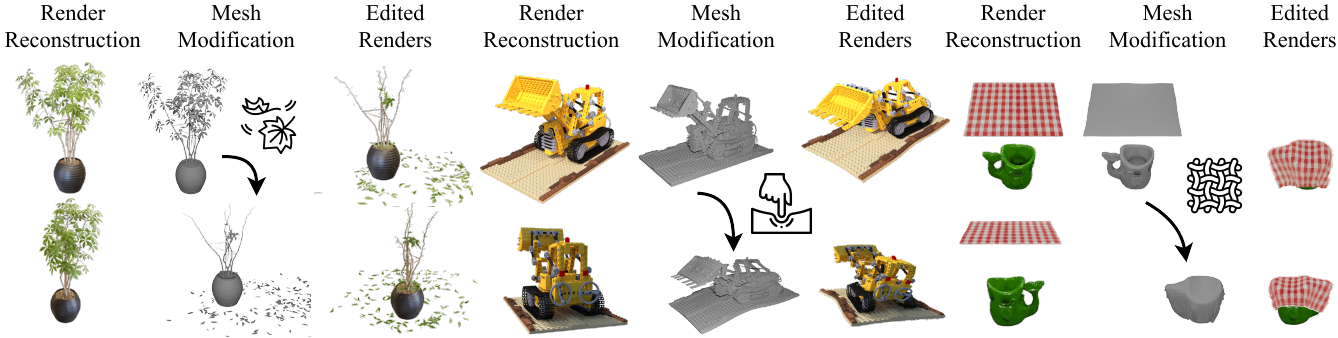} 
            \caption{\textbf{Physical simulations.} From left to right: (1) Rigid body simulation of falling leaves. (2) Soft body simulation of the Lego dozer being squished. (3) Cloth simulation of fabric falling onto a cup. The middle columns show the deformation-driving meshes.
            }
            \label{fig: symulations_synthetic}
            \vspace{-0.2cm}
        \end{figure*}

        Recent years have seen rapid progress in 3D scene representation and rendering, driven by applications in robotics, virtual environments, and content creation that increasingly demand physically grounded simulation and interactive editing \citep{ wang2023seal3d, Genesis, huang2024blenderalchemy}. Tasks such as object manipulation, deformable modelling, collision handling, and physics-aware animation require 3D representations that are both high-fidelity and intuitively editable while remaining compatible with physics engines. 
        
        Neural Radiance Fields (NeRFs) \citep{mildenhall2020nerf} achieve high visual fidelity by modelling scenes as continuous volumetric functions capable of high-quality novel-view synthesis and complex view-dependent effects. However, NeRFs encode spatial structure implicitly within network parameters, making localized scene edits difficult to perform without retraining \citep{wang2023seal3d, weber2024nerfiller}. This limitation restricts their applicability in interactive and physically grounded settings. Several works seek to mitigate this issue by introducing explicit control structures, including point-based conditioning \citep{wang2023rip, chen2023neuraleditor, zhang2023papr}, mesh-based control \citep{yuan2022nerf, yang2022neumesh}, or primitive-based representations for simulation \citep{monnier2023differentiable}. While these approaches enable limited forms of manual editing, they are typically constrained to coarse modifications and often introduce visual artifacts.
        Recent advances in explicitly parametrized scene representations demonstrate that spatial locality and explicit structure can substantially improve editability and interaction \citep{kerbl20233d, malarz2025gaussian, borycki2024gasp}. These results highlight desirable properties for editable 3D representations, but do not directly address how such locality can be integrated into NeRF models.
        Motivated by these observations, we address a fundamental limitation in NeRF editing task: the absence of a transformation-aware space encoding.
        Existing NeRF encodings, including positional encodings \citep{mildenhall2020nerf} and multi-resolution hash grids \citep{muller2022instant}, entangle features globally across space, causing localized modifications to propagate undesirably and preventing precise control. Prior attempts to alleviate this issue rely on point-based latent representations that store features at discrete spatial locations and interpolate them at query points \citep{xu2022point, chen2023neuraleditor, wang2023rip}. While this enables explicit spatial manipulation, such representations remain highly sensitive to deformations, as changes in point positions alter local neighbourhood and lead to unstable interpolation (see Fig. \ref{fig:knn_comparisons}).
        
        In this work, we introduce Affine-\textbf{E}quivariant \textbf{K}ernel \textbf{S}pace Encoding (\our{}), a novel positional encoding mechanism for NeRFs. Our approach represents scene features in a continuous kernel space, where each latent element defines a spatially localized and anisotropic region of influence. This formulation replaces discrete point samples and grid-based embeddings with a variant kernel field that supports stable and localized feature evaluation.
        In practice, kernels are parametrized as anisotropic Gaussians, enabling efficient feature evaluation via a $k$-nearest neighbour search weighted by the Mahalanobis distance. By accounting for local anisotropy through kernel covariances, this interpolation remains stable under spatial transformations while capturing richer local geometric structure than point-based representations.
        To retain the expressiveness of multi-resolution encodings, we further propose a training-time feature distillation mechanism that transfers spatial detail from hash grid encodings into the kernel field. Unlike prior approaches that embed Gaussian features within fixed grids \citep{govindarajan2024lagrangian}, the resulting representation is fully decoupled from grid structures at inference time, yielding an editable, deformable, and compact latent field that preserves high reconstruction quality.

\section{Related Works}

        \begin{figure*}[t]
            \centering
            \includegraphics[width=\textwidth]{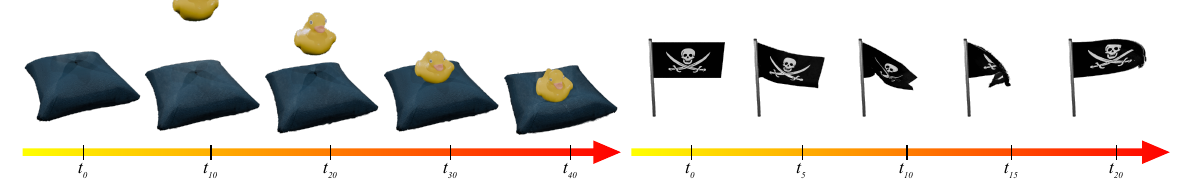}
            \caption{\textbf{Evolution of two physical simulations.} From left to right: (1) A rubber duck falling onto a pillow and deforming it. (2) A pirate flag waving under the influence of wind. Both simulations are performed on our own assets.}
            \label{fig: symulations_timeline}
            \vspace{-0.3cm}
        \end{figure*}

        Several approaches focus on modeling deformation or displacement fields at a per-frame level~\citep{park2021nerfies,park2021hypernerf,tretschk2021non,weng2022humannerf}, while others aim to capture continuous motion over time by learning time-dependent 3D flow fields~\citep{du2021neural,gao2021dynamic,guo2023forward,cao2023hexplane}.
        
        A substantial body of research has also explored NeRF-based scene editing across different application domains. This includes methods driven by semantic segmentation or labels~\citep{bao2023sine,dong2023vica,haque2023instruct,mikaeili2023sked,song2023blending,wang2022clip}, as well as techniques that enable relighting and texture modification through shading cues~\citep{gong2023recolornerf,liu2021editing,rudnev2022nerf,srinivasan2021nerv}. Other efforts support structural changes in the scene, such as inserting or removing objects~\citep{kobayashi2022decomposing,lazova2023control,weder2023removing}, while some are tailored specifically for facial editing~\citep{hwang2023faceclipnerf,jiang2022nerffaceediting,sun2022fenerf} or physics-based manipulation from video sequences~\citep{hofherr2023neural,qiao2022neuphysics}
        Geometry editing within the NeRF framework has received considerable attention~\citep{kania2022conerf,yuan2023interactive,zheng2023editablenerf}.
        
        Our model uses geometry editing and physics simulations.
        Existing methods leverage various geometric primitives for conditioning NeRFs, most notably 3D point clouds. For instance, RIP-NeRF~\citep{wang2023rip} introduces a rotation-invariant point-based representation that enables fine-grained editing and cross-scene compositing by decoupling the neural field from explicit geometry. NeuralEditor~\citep{chen2023neuraleditor} adopts a point cloud as the structural backbone and proposes a voxel-guided rendering scheme to facilitate precise shape deformation and scene morphing. Similarly, PAPR~\citep{zhang2023papr} learns a parsimonious set of scene-representative points enriched with learned features and influence scores, enabling geometry editing and appearance manipulation.
        
        Some approaches leverage explicit mesh representations to enable NeRF editing. NeRF-Editing~\citep{yuan2022nerf} extracts a mesh from the scene and allows users to apply traditional mesh deformations, which are then transferred to the implicit radiance field by bending camera rays through a proxy tetrahedral mesh. Similarly, NeuMesh~\citep{yang2022neumesh} encodes disentangled geometry and texture features at mesh vertices, enabling mesh-guided geometry editing.
        To reduce computational complexity, some approaches rely on simplified geometry proxies, such as coarse meshes paired with cage-based deformation techniques~\citep{jambon2023nerfshop, peng2022cagenerf, xu2022deforming}. VolTeMorph~\citep{garbin2024voltemorph} introduces an explicit volume deformation technique that supports realistic extrapolation and can be edited using standard software, enabling applications such as physics-based object deformation and avatar animation. PIE-NeRF~\citep{feng2024pie} integrates physics-based, meshless simulations directly with NeRF representations, enabling interactive and realistic animations.
        
        While existing approaches enable manual editing via explicit conditioning representations, they often rely on complex, task-specific pipelines~\citep{feng2024pie, jambon2023nerfshop, garbin2024voltemorph}. 
        Point-conditioning methods \citep{wang2023rip, xu2022point, chen2023neuraleditor} alleviate this issue and provide localized control, but typically exhibit limited edit quality in contrast to \our{}.

\section{Preliminary}

        Our method, \our{}, is formulated within the Neural Radiance Field framework and introduces a kernel-based latent space encoding inspired by multi-resolution hash grids and Gaussian kernel representations. In this section, we briefly review the relevant background on neural radiance fields and spatial encoding methods.
    
    \paragraph{Neural Radiance Fields}
    
        Vanilla NeRF~\citep{mildenhall2020nerf} represents a 3D scene as a continuous volumetric field by learning a function that maps a spatial location $\mathbf{x} = (x, y, z)$ and a viewing direction $\mathbf{d} = (\theta, \psi)$, to an emitted colour $\mathbf{c} = (r, g, b)$ and a volume density $\sigma$. Formally, the scene is approximated by a multi-layer perceptron (MLP):
        \begin{equation}
        \mathcal{F}_{\text{NeRF}}(\mathbf{x}, \mathbf{d}; \Theta) = (\mathbf{c}, \sigma),
        \end{equation}
        where $\Theta$ denotes the trainable network parameters.
        
        The model is trained using a set of posed images by casting rays from each camera pixel into the scene and accumulating colour and opacity along each ray based on volumetric rendering principles. The goal is to minimize the difference between the rendered and ground-truth images, allowing the MLP to implicitly encode both the geometry and appearance of the 3D scene.

        \begin{figure*}[t]
        \centering
        \includegraphics[width=\textwidth]{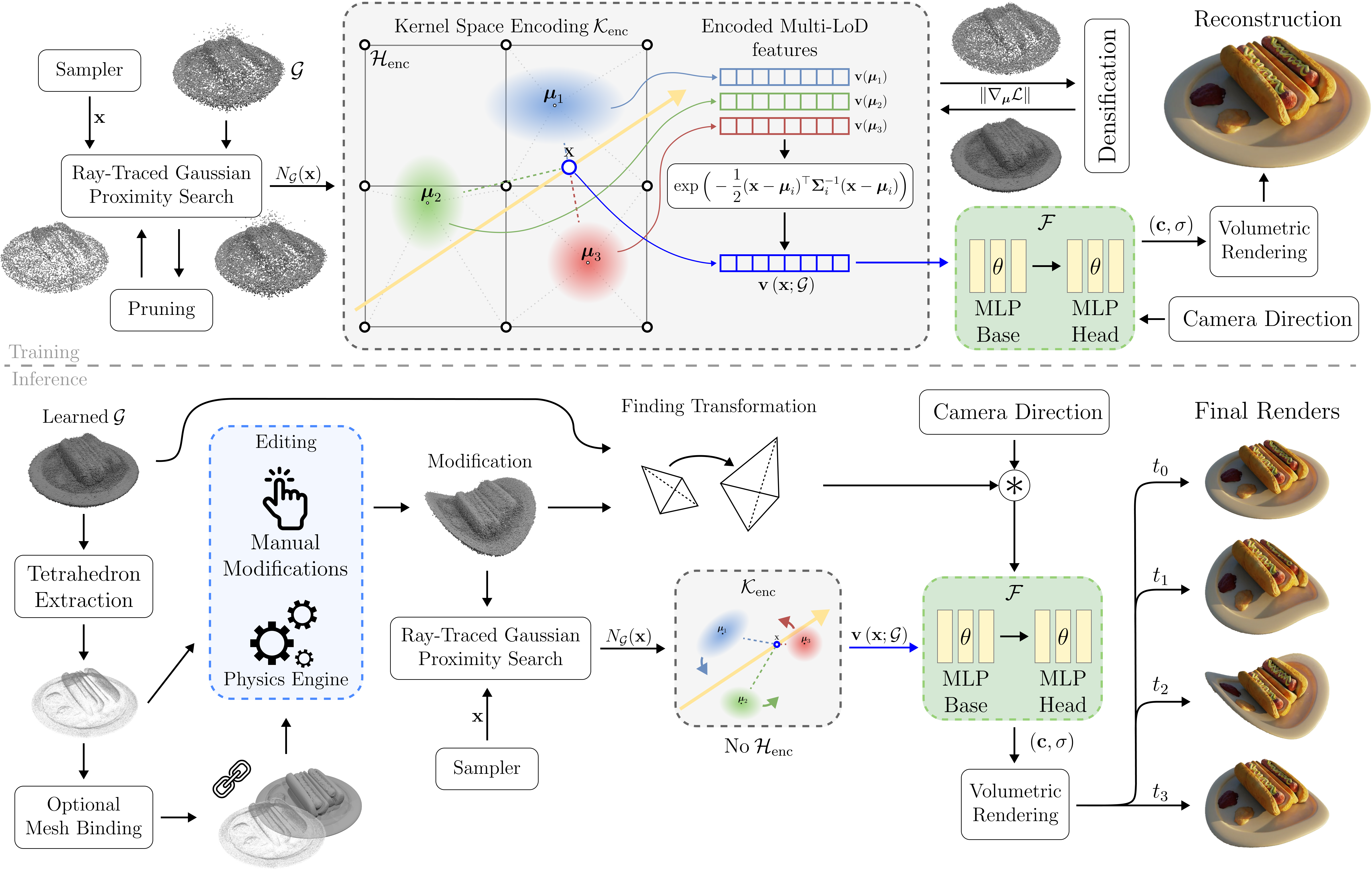} 
        \caption{\textbf{Model overview.}
        Top: During training, a subset of Gaussians is selected using \ourknn{} (\ourknnshort{}), which also handles pruning. The nearest Gaussians to the sampling position $\mathbf{x}$ are passed to the \ourencoding{}, which interpolates their features to produce the final positional embedding $\mathbf{v}(\mathbf{x}; \mathcal{G})$. The embedding is then processed by the neural network $\mathcal{F}$ to predict colour $\mathbf{c}$ and opacity $\sigma$, which are used for volumetric rendering.
        Bottom: At inference time, the learned Gaussians serve as input parameters and can undergo manual or physics-driven edits. The edited Gaussians are passed through the same rendering pipeline to generate the final image, with the view-direction input to $\mathcal{F}$ adjusted by the inverse rotation of the modified Gaussians. Since the kernel space encoding is fixed after training, the auxiliary network $\mathcal{H}_{\text{enc}}$ is omitted during inference.
        } 
        \label{fig:model} 
        \end{figure*}
        
    \paragraph{Hash Grid Encoding}
    
         Many NeRF variants adopt the Hash Grid Encoding ~\citep{muller2022instant}, to improve scalability and spatial precision which captures high-frequency scene details by dividing space into multiple Levels of Detail (LoD), each with trainable parameters $\Phi$ and feature vectors $F$. These levels vary in resolution, allowing the encoding to represent both coarse and fine details. For a query point $\mathbf{x}$, the output feature vector $\mathbf{v}$ is obtained by concatenating trilinearly interpolated features from all levels, based on $\mathbf{x}$'s position within the grid

        \begin{equation}
        \mathcal{H}_{\text{enc}}(\mathbf{x}; \Phi) = \mathbf{v}(\mathbf{x}).
        \end{equation}
    
        \paragraph{Gaussian Kernels}

            Gaussian kernels define smooth, spatially localized basis functions in $\mathbb{R}^3$ with anisotropic support, making them well suited for continuous spatial representations and deformation-aware interpolation. Since they preserve the neighbourhood during affine transformations.

            We denote a set of Gaussian kernels as
            \begin{equation}
            \left\{ \mathcal{N}(\boldsymbol{\mu}_i, \mathbf{\Sigma}_i) \right\}_{i=1}^n.
            \end{equation}

\section{Proposed Method}    

        Our method, called \our{}, integrates affine-equvariant transformation properties of Gaussian kernels and a neural network-based rendering procedure into a single system. Specifically, we use a set of Gaussian kernels, enhanced with a trainable latent feature vector $\mathbf{v} \in \mathbb{R}^n$. We refer to this set of Gaussians as $\mathcal{G}$. 
        
        We use a NeRF-based neural network $\mathcal{F}$ to predict colour and opacity from the nearest Gaussian features. Formally, the model is defined as:

        \begin{equation}
        \mathcal{F}(\mathbf{x}, \mathbf{d}; \mathcal{G}, \Theta) = (\mathbf{c}, \sigma),
        \end{equation}
        
        where $\Theta$ denote the trainable network parameters. The model, alongside the standard NeRF input, takes a set of trainable Gaussians $\mathcal{G}$ and outputs colour $\mathbf{c}$ and density $\sigma$ at any query point, enabling neural rendering conditioned on nearby Gaussian features.

        \begin{figure}[h!]
        \centering
        \includegraphics[width=\linewidth]{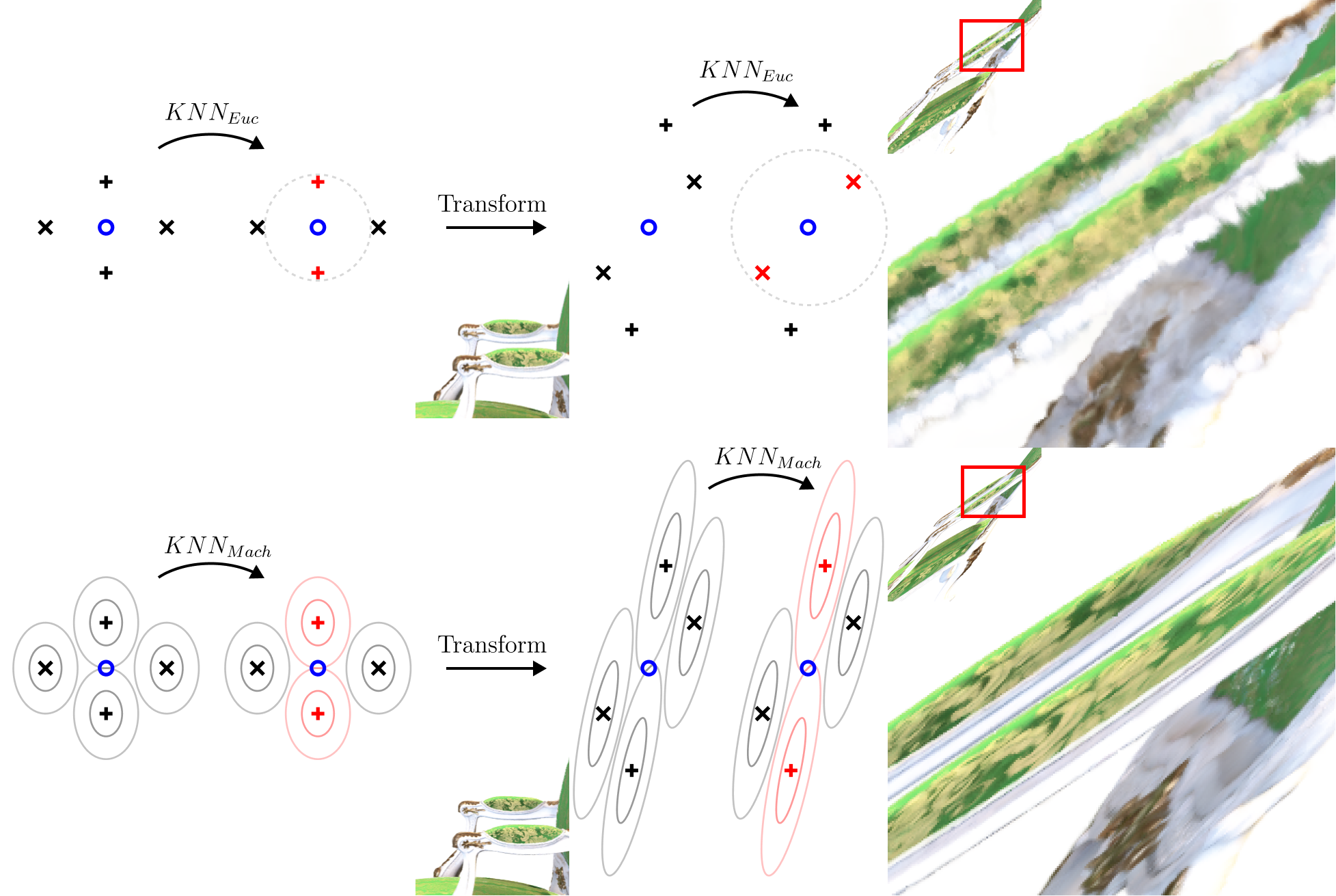}
        \caption{\textbf{KNN Comparisons. } Comparison of neighbourhood changes under deformation using Euclidean distance KNN (top) versus our proposed Mahalanobis distance KNN (bottom). Moving points in traditional encodings changes local neighbourhoods inconsistently, causing unstable feature interpolation. Our method preserves relative feature structure under spatial transformations and yields visibly improved results with no holes and distortions.
        }
        \label{fig:knn_comparisons}
        \end{figure}

    \paragraph{Kernel Space Encoding}
        In point-based encodings, local neighbourhoods change inconsistently under spatial deformations, leading to unstable interpolation (see Fig.~\ref{fig:knn_comparisons}). Our encoding resolves this by representing latent features with anisotropic Gaussian kernels and using a Mahalanobis-distance-based interpolation that respects local geometry. Our encoding takes a set of query points $\mathbf{x}$ as input and a set of learnable Gaussians parameters $\mathcal{G}$, producing multi-resolution features. Formally, we define this encoding as:
        
        \begin{equation}
        \mathcal{K}_{\text{enc}}\left(\mathbf{x}; \mathcal{G}\right) = \mathbf{v}(\mathbf{x})
        \end{equation}
        
        Unlike the traditional \textit{Hash Grid Encoding}, where the output depends directly on the query point $\mathbf{x}$, here the features are derived from nearby Gaussians. We select the $N$ closest Gaussians to $\mathbf{x}$ using our \ourknnshort{} algorithm (detailed in the following section). The final feature vector is computed as a weighted interpolation of the Gaussian features using a Mahalanobis-distance-based weighting scheme:
                
        \begin{equation}
        \mathbf{v}\left(\mathcal{G}\right) = \sum_{i=1}^{k} w_i(\mathbf{x}, \mathcal{G}) \cdot \mathbf{v}_i,
        \label{eq: gaussian_feature}
        \end{equation}
        
        \begin{equation}
        w_i(\mathbf{\mathbf{x}, \mathcal{G}}) = \exp\Big(-\frac{1}{2}(\mathbf{x} - \boldsymbol{\mu}_i)^\top \boldsymbol{\Sigma}_i^{-1} (\mathbf{x} - \boldsymbol{\mu}_i)\Big),
        \end{equation}

        where $w_i(\mathbf{\mathbf{x}, \mathcal{G}})$ is the interpolation weight, $k$ is the number of nearest neighbours considered, and $\boldsymbol{\Sigma}_i$ is the full anisotropic covariance of the $i$-th Gaussian kernel.
        
    \paragraph{\ourknn{}}

\begin{figure}[t]
        \centering
        \includegraphics[width=0.8\columnwidth]{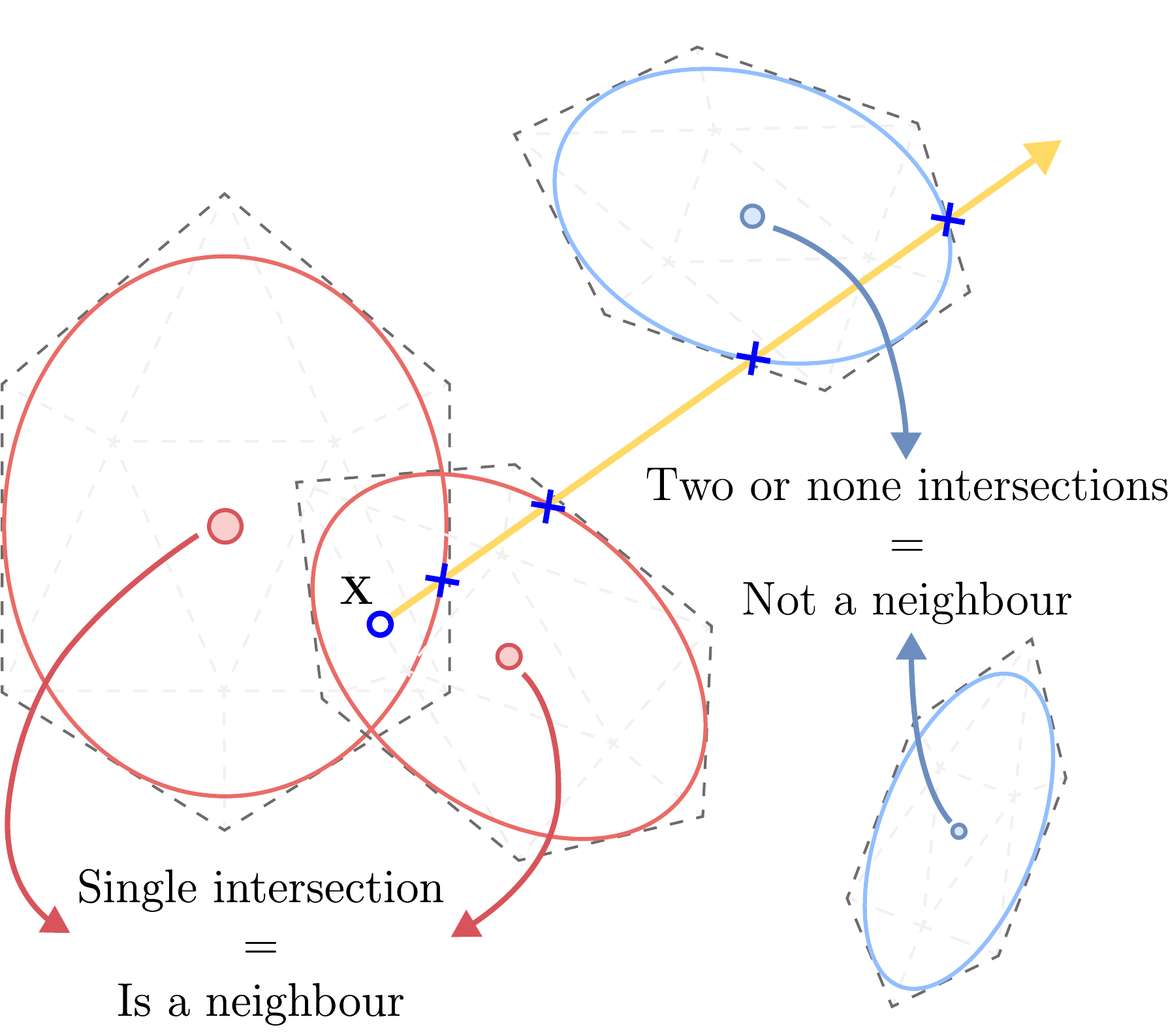}
        \caption{\textbf{The \ourknnshort{} working principle.} A light ray passing through the scene is illustrated, along with its intersections with the icosahedrons. The figure highlights which Gaussians are considered neighbors and which are not.}
        \label{fig:knn}
        \end{figure}

        To achieve affine transformation equivariance, nearest-neighbour search around a query point must be performed using the Mahalanobis distance. To make this process efficient, we restrict nearest-neighbour candidates to Gaussians whose confidence ellipsoids (defined by a quantile parameter $Q$) contain the query point $\mathbf{x}$. This reduces the neighbour search to a point-in-ellipsoid test, which we approximate using circumscribed stretched icosahedra. This approach extends the RT-kNNS algorithm~\citep{Nagarajan2023RTKNNS}. Unlike RT-kNNS, \ourknnshort{} performs the point-in-ellipsoid test individually for each Gaussian, where the Gaussian mean corresponds to a KNN candidate for the query point $\mathbf{x}$. Following~\citep{Nagarajan2023RTKNNS}, we trace rays originating from $\mathbf{x}$ and collect Gaussians whose confidence ellipsoids produce exactly one ray–ellipsoid intersection (see Fig.~\ref{fig:knn}). A sorted hit buffer maintains up to $k$ nearest-neighbour candidates based on the squared Mahalanobis distance to $\mathbf{x}$. 

        \paragraph{Hash Grid Feature Distillation} 
        While hash-grid encodings are effective for representing static scenes, they do not support precise, localized edits. Modifying vertices at lower levels of detail propagates changes to all features within the corresponding voxel, often affecting higher-resolution details and producing inconsistent, unintuitive results. To address this limitation, we introduce a Hash Grid Feature Distillation mechanism, which decouples the feature representation from the underlying grid vertices and transfers it to a set of Gaussian kernels. During training, both the hash-grid parameters $\Phi$ and the Gaussian positions $\boldsymbol{\mu}_i$ are optimized jointly, allowing the Gaussians to explore the multi-resolution feature space and shape the latent encoding. The Gaussian features $\mathbf{v}(\mathbf{x})$ are sampled from the hash-grid encoding at the kernel centres, formally described as:

        \begin{equation}
        \mathbf{v}\left(\mathbf{x}\right) = \sum_{i=1}^{k} w_i(\mathbf{x}, \mathcal{G}) \cdot \mathcal{H}_{\text{enc}}(\boldsymbol{\mu}_i; \Phi),
        \end{equation}
                
        At inference, we fall back to the equation \ref{eq: gaussian_feature} the hash grid is no longer needed. The Gaussians retain their learned feature vectors, which remain fixed. Since interpolation operates solely over these Gaussian features, any adjustments to Gaussian positions, rotations, or scales directly modify the rendered output.
        
    \paragraph{View-Direction Restoration}
    
        After deformation, some Gaussians may be observed from previously unseen directions. To maintain consistent appearance, we need to restore their view-dependent features as if no deformation had occurred. \citet{chen2023neuraleditor} addressed this by assigning a separate local coordinate system to each point in space and tracking its transformation during deformation, which increases representation size.
        In contrast, our Gaussians already have principal axes that define local coordinate systems. This naturally allows us to track their spatial transformations efficiently using the Kabsch algorithm. By monitoring these axes, we can both restore view-dependent features and update the anisotropic scales of the Gaussians consistently after deformation.
        
    \paragraph{Pruning and Densification}
        To enable Gaussian kernels to better represent the latent feature space, we adopt densification and pruning strategies that regulate the number of Gaussians during training. For densification, we follow the approach of \citep{kerbl20233d}, tracking Gaussian means via their gradients and cloning or splitting Gaussians accordingly. Unlike \citep{xu2022point}, we initialize the features of new Gaussians by sampling from the hash-grid encoding rather than from nearby shading information, ensuring better alignment with the latent feature field.
        For pruning, we track which Gaussians are actively selected as neighbours by our \ourknnshort{} algorithm. Gaussians that are not used as neighbours for several consecutive iterations are removed, resulting in a more compact and efficient representation.

    \paragraph{Editing}

        Thanks to the \our{} feature encoding, the latent space is structured around the spatial configuration of the Gaussian kernels. This alignment allows direct edits in the coordinate space of the Gaussians, effectively translating spatial transformations into consistent latent-space manipulations. By weighting feature interpolation according to the Mahalanobis distance, we maintain affine transformation equivariance and retain the local density structure of the underlying Gaussians. As a result, the latent features remain coherent after deformation, ensuring that modifications produce smooth, stable, and physically consistent updates in the rendered scene without requiring network retraining.

        In practice, for the editing task, we export Gaussians as tetrahedra, where each orthogonal arm corresponds to a principal axis of the Gaussian. This representation allows us to explicitly track how the scale and rotation of each Gaussian are affected by an edit, and additionally provides the information required for view-direction restoration. Edits can be applied directly to the tetrahedra or, alternatively, the tetrahedra can be bound to a mesh for intuitive manipulation. After editing, the modified tetrahedra are converted back into Gaussians and used as parameters for the kernel space encoding. Interpolation between these modified Gaussians then enables the system to synthesize novel views of the edited scene.
    
\section{Experiments}

    We design our experiments to demonstrate that \our{} maintains the reconstruction quality of state-of-the-art (SOTA) methods while enabling complex object modifications.

    \begin{figure}[t]
    \centering
    \includegraphics[width=\columnwidth]{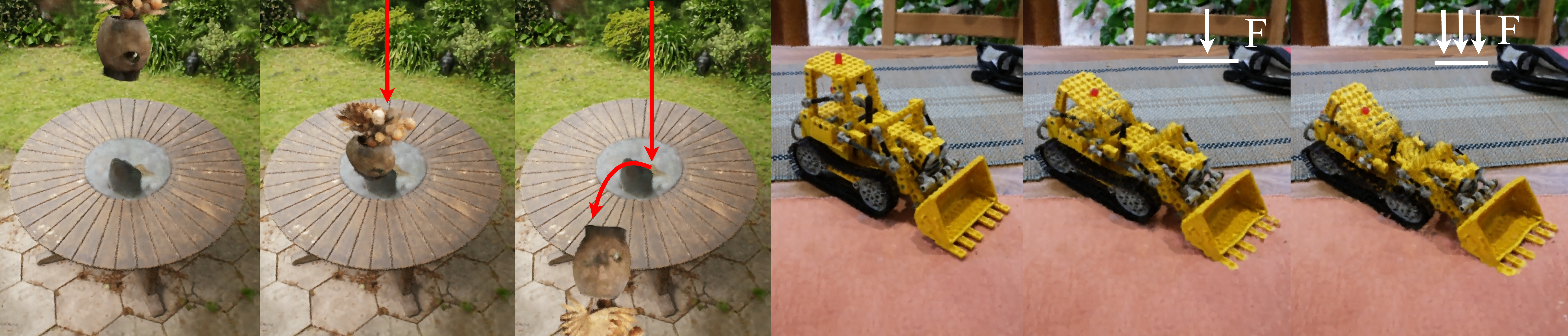}
    \caption{\textbf{Example edits on real-world scenes.} From left to right: (1) Physics-based simulation, showing an object falling onto a tilted table and bouncing off. (2) Physics simulation, where a force is applied to deform a plasticine dozer.}
    \label{fig: symulations_real}
    \vspace{-0.4cm}
    \end{figure}
    
        \paragraph{Datasets}

            Following prior work, we evaluate on the \textit{NeRF-Synthetic} dataset~\citep{mildenhall2020nerf}, which contains eight synthetic scenes with diverse geometry, texture, and specular properties. Additionally to synthetic data we trained our NeRF model trained on the \textit{Mip-NeRF 360} dataset~\citep{barron2022mipnerf360}, comprising five outdoor and four indoor real-world 360\textdegree scenes. To further demonstrate editing capabilities, we include the \textit{fox} scene from Instant-NGP~\citep{muller2022instant}, and introduce a custom set of 3D assets with deformable and articulated objects, enabling dynamic scene editing and physical interaction.

        \paragraph{Baselines}

            We compare \our{} against both static and editable point-based and Gaussian-based scene representations. For static radiance field models, we evaluate Instant-NGP \citep{muller2022instant}, which introduced the hash-grid encoding and serves as the foundation of our neural field, as well as LagHash \citep{govindarajan2024lagrangian}, which augments hash-grid encodings with Gaussian primitives. While both methods achieve high reconstruction quality, they do not support scene editing.
            
\begin{figure*}[t]
            \centering
            \includegraphics[width=\textwidth]{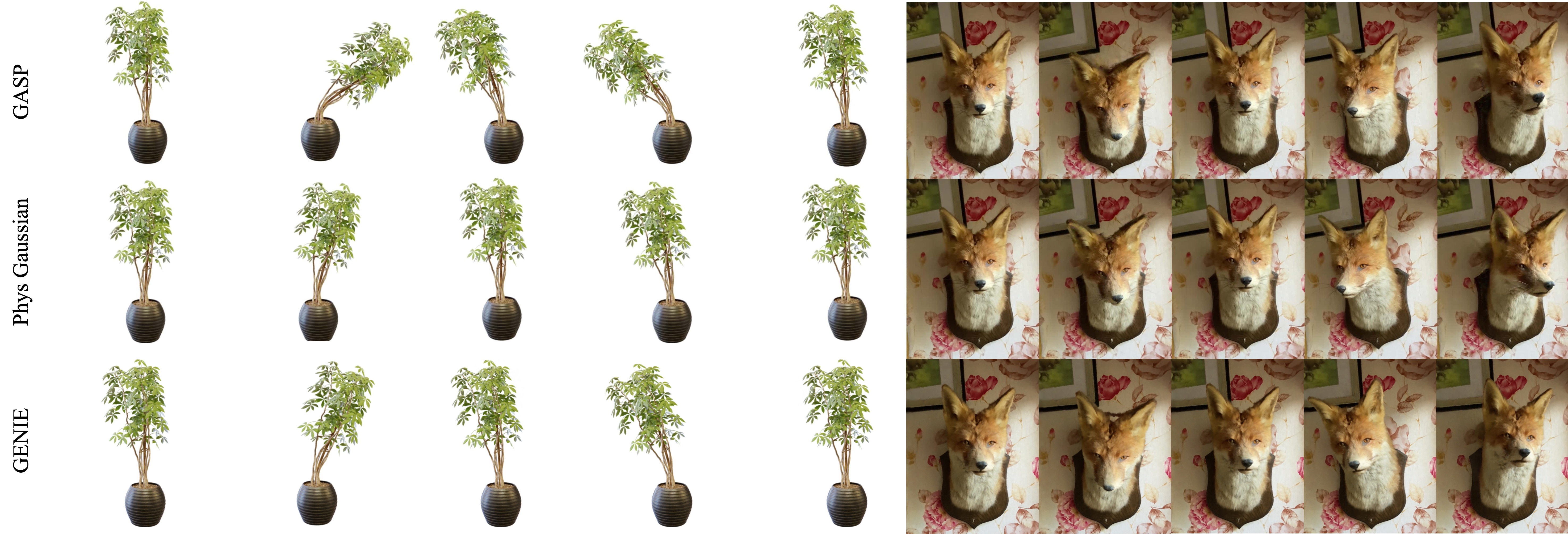}
            \caption{\textbf{Physics simulations with Gaussian Splatting methods.} From left to right: (1) wind simulation on the ficus plant from, (2) particle impact simulation on the fox head. Results shown for PhysGaussian \citep{xie2024physgaussian} and GASP \citep{borycki2024gasp}.}
            \label{fig: phys_sim_comparison}
            \vspace{-0.2cm}
            \end{figure*}

            For editable representations, we compare against RIP-NeRF \citep{wang2023rip}, Point-NeRF \citep{xu2022point}, and Neuraleditor \citep{chen2023neuraleditor}, which enable scene editing using point-based NeRF formulations. We additionally include a naive plotting baseline \cite{chen2023neuraleditor} that renders a transformed dense point cloud by directly projecting points onto the camera plane using per-point opacity and view-dependent color.
            These baselines are selected to demonstrate that \our{} not only achieves reconstruction quality comparable to or exceeding SOTA methods, while enabling editing with significantly fewer artifacts. Furthermore, we provide qualitative comparisons of physics-based simulations against PhysGaussian \citep{xie2024physgaussian} and GASP \citep{borycki2024gasp}, two Gaussian-based methods designed for physical interaction, as shown in Fig.~\ref{fig: phys_sim_comparison}.
    

        \paragraph{Quantitative Results}

            \begin{table}[h!]
            {\small 
            \begin{center}
                \begin{tabular}{l@{\;}l@{\;}l@{\;}l@{\;}l@{\;}l@{\;}l@{\;}l@{\;}l@{\;}l@{}}
                     & \rot{Chair} & \rot{Drums} & \rot{Lego} & \rot{Mic} & \rot{Materials} & \rot{Ship} & \rot{Hotdog} & \rot{Ficus} \\ 
             \hline
            
                \multicolumn{9}{c}{Non Editable} \\ 
             \hline

            INGP & 31.97 & 22.67 & 33.44 & 31.38 & 22.66 & 28.83 & 34.04 & 29.47 \\
            LagHash & \bf 35.66. & \bf 25.68 & \bf 35.49 & \bf 36.71 & \bf 29.60 & \bf 30.88 & \bf 37.30 & \bf 33.83 \\
            
            \hline
                \multicolumn{9}{c}{Editable} \\ 
            \hline
            GaMeS & \bf 35.73 & 26.15  & 35.57  & 35.67  & 29.89  & 30.78  & \bf 37.58  & 34.83 \\
            RIP-NeRF & 34.84 & 24.89 & 33.41 & 34.19 &28.31 & 30.65 & 35.96 & 32.23 \\
            Point-NeRF & 35.40 & 26.06 & 35.04 & 35.95 & 29.61 & 30.97 & 37.30 & \bf 36.13 \\
            Neuraleditor & 34.94 & \bf 26.19 & 34.28 & 36.09 & \bf 30.38 & 29.99 & 36.70 & 33.64 \\
            \our{} & 34.72 & 26.01 & \bf 35.59 & \bf 36.54 & 30.08 & \bf 31.10 & 37.11 & 33.82 \\
            \hline

                \end{tabular}
                \end{center}
                }
                \caption{Quantitative comparisons (PSNR) on a NeRF-Synthetic dataset showing that \our{} gives comparable results with other models on static scenes.}
                \vspace{-0.3cm}
                \label{tab:nerf_synthetic}
            \end{table}

            We present quantitative results on the \textit{NeRF-Synthetic} dataset in two settings: reconstruction of static scenes (Table~\ref{tab:nerf_synthetic}) and reconstruction after edits (Table~\ref{tab:editing_results}). 
            For static scene reconstruction, \our{} achieves quality comparable to state-of-the-art editable methods, and in some cases provides the best results among methods that support editing. This demonstrates that our approach preserves rendering quality while enabling scene edits.
            
            For edited scene reconstruction, we evaluate on the editing benchmark introduced by \citet{chen2023neuraleditor}, which applies handcrafted modifications to the \textit{NeRF-Synthetic} dataset and provides ground-truth edited images. As shown in Table~\ref{tab:editing_results}, our method consistently outperforms prior state-of-the-art approaches, including GaMeS, a purely Gaussian Splatting–based editing method.

            \begin{table}[h!]
            \vspace{-0.1cm}
            {\small 
            \begin{center}
                \begin{tabular}{@{}l@{\;}l@{\;}l@{\;}l@{\;}l@{\;}l@{\;}l@{\;}l@{\;}l@{\;}}
                     & \rot{Chair} & \rot{Drums} & \rot{Lego} & \rot{Mic} & \rot{Materials} & \rot{Ship} & \rot{Hotdog} & \rot{Ficus} \\ 
            \hline
            \multicolumn{9}{c}{ PSNR } \\ 
            \hline
            
            Naive Plotting & 24.58 & 21.54 & 25.38 & 27.56 & 21.59 & 22.21 & 26.72 & 24.62 \\ 
            Neuraleditor & 25.29 & 21.93 & 27.14 & 27.49 & 23.04 & 24.12 & 27.14 & 24.83 \\ 
            GaMeS & 24.51 & 22.02 & 26.65 & 27.07 & 21.73 & 22.19 & 27.26 & 26.65 \\ 
            \our{} & \bf 26.03 & \bf 22.08 & \bf 28.04 & \bf 27.85 & \bf 23.14 & \bf 24.43 & \bf 28.23 & \bf 27.58 \\
            
            \hline
            \multicolumn{9}{c}{ SSIM } \\ 
            \hline
            
            Naive Plotting & 0.930 & 0.892 & 0.904 & 0.956 & 0.867 & 0.807 & 0.930 & 0.925 \\
            Neuraleditor & 0.944 & 0.900 & 0.945 & 0.958 & 0.887 & 0.832 & 0.937 & 0.927 \\
            GaMeS & 0.941 & \bf 0.914 & 0.936 & 0.960 & 0.890 & 0.811 & 0.947 & 0.947 \\
            \our{} & \bf 0.957 & 0.910 & \bf 0.961 & \bf 0.964 & \bf 0.911 & \bf 0.855 & \bf 0.962 & \bf 0.951 \\
            \hline

            \multicolumn{9}{c}{ LPIPS } \\ 
            \hline
            
            Naive Plotting & 0.050 & 0.107 & 0.066 & 0.053 & 0.126 & 0.187 & 0.085 & 0.072 \\
            Neuraleditor & 0.041 & 0.100 & 0.038 & 0.050 & 0.103 & 0.158 & 0.078 & 0.069 \\
            GaMeS & 0.039 & \bf 0.067 & 0.035 & \bf 0.032 & 0.077 & 0.177 & 0.046 & \bf 0.036 \\
            \our{} & \bf 0.030 & 0.071 & \bf 0.023 & 0.036 & \bf 0.062 & \bf 0.143 & \bf 0.037 & \bf 0.036 \\
            \hline

                \end{tabular}
                \end{center}
                }
                \caption{Quantitative comparisons (PSNR) on a \cite{chen2023neuraleditor} benchmark showing that \our{} achieves best results in editing task.}
                \label{tab:editing_results}
                \vspace{-0.3cm}
            \end{table}

        \paragraph{Qualitative Results}

            \begin{figure}[t]
            \centering
            \includegraphics[width=\columnwidth]{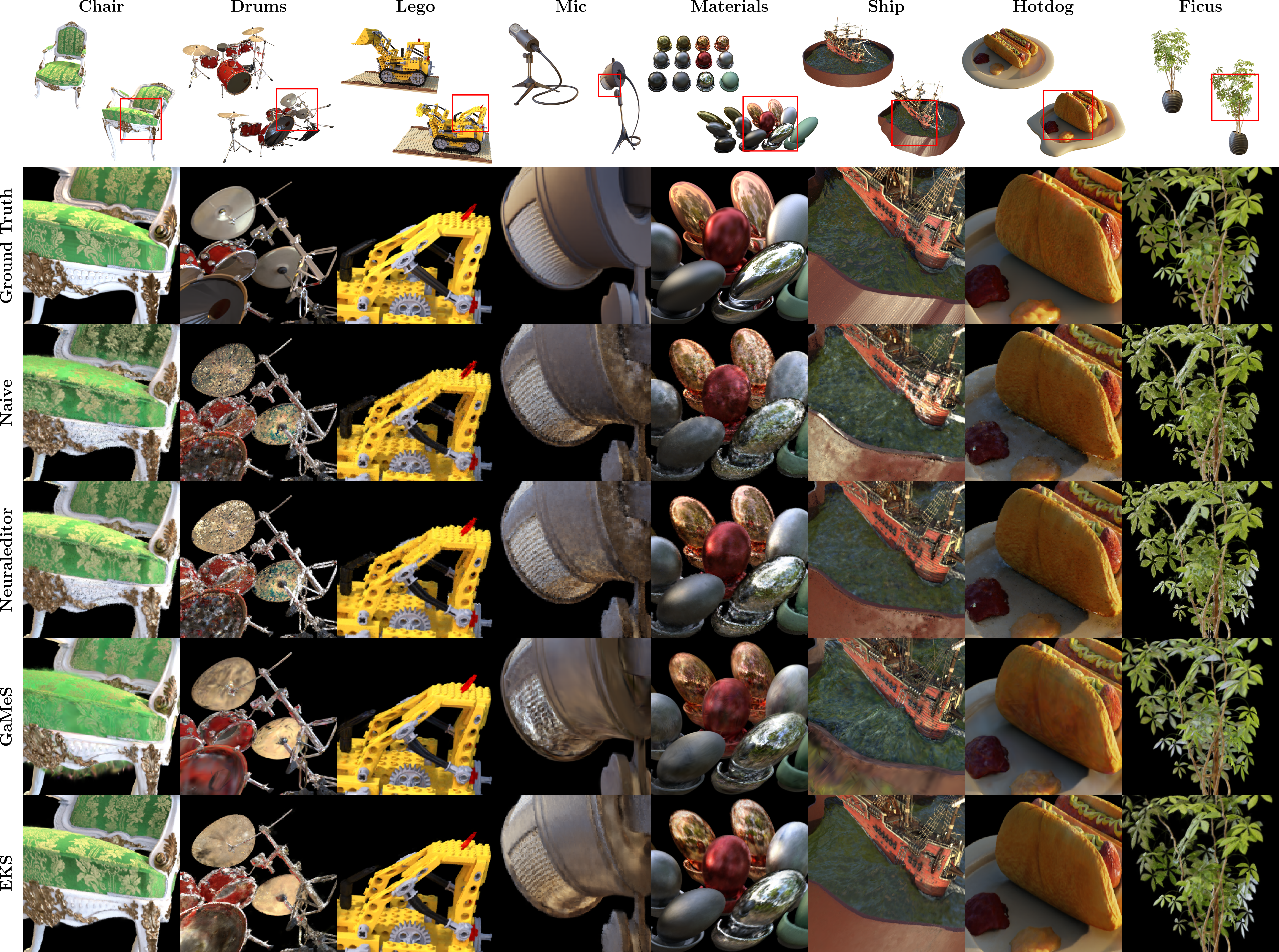}
            \caption{\textbf{Qualitative comparison.} Results shown on the \textit{NeRF-Synthetic} dataset. Modified objects are in the top row. Each row compares reconstruction quality across different methods. Enlarged version is presented in Appendix~\ref{adx:enlarged}}
            \label{fig: qualitative}
            \vspace{-0.5cm}
            \end{figure} 
            
            For qualitative evaluation, we use the editing benchmark of \citet{chen2023neuraleditor} and assess the visual quality of edits across methods. We observe that \our{} produces higher-quality results in the zero-shot editing setting. In particular, it better preserves fine details while yielding noticeably smoother flat surfaces. In the \textit{Drums} scene, the gong is consistently restored without visible holes. In contrast, Neuraleditor and other point-based methods exhibit visible granularity across the image in all cases, and their edits are sometimes inconsistent, leaving holes in the reconstructed scenes. Additional artifacts are also observed, such as distortions on the plate in the \textit{Hotdog} scene.
            In the Gaussian Splatting–based method GaMeS, we observe a different class of artifacts: individual Gaussians remain visible beneath the \textit{Chair}, and in the \textit{Ship} scene the Gaussians appear excessively scaled, causing them to bleed outside the bowl geometry. Additionally, Gaussian primitives remain visible across several scenes. In contrast, \our{} avoids these artifacts and consistently produces smooth surfaces with high reconstruction quality across all evaluated scenes, as the Gaussians encode latent features rather than explicit geometry and are accessed only through local KNN-based interpolation.
                    
        \paragraph{Physic-based Editing}

            We conducted a series of physics-based simulations in Blender \citep{blender} using the mesh-driven editing mechanism described earlier. These experiments span both synthetic and real-world datasets and include diverse physical phenomena such as rigid body dynamics, soft body deformation, and cloth simulation. In these scenarios, deformations of the driving mesh were used to update the corresponding Gaussian components in real time, enabling seamless integration of physical interactions into the scene. In addition, we performed simulations following PhysGaussian \citep{xie2024physgaussian} and compared \our{} qualitatively against both PhysGaussian and GASP \citep{borycki2024gasp}.        
            
            The results of these simulations are illustrated in Figs~\ref{fig: symulations_timeline}, \ref{fig: symulations_synthetic}, \ref{fig: symulations_real}, and \ref{fig: phys_sim_comparison}. These visualizations demonstrate that \our{} produces realistic and physically plausible edits across a wide range of scenarios. Whether simulating leaves falling from a plant, squashing a soft object, or draping cloth over complex geometry, our method maintains high rendering fidelity while enabling expressive and controllable scene manipulation. This highlights the potential of \our{} as a flexible framework for neural scene editing driven by physical interactions.

        \paragraph{Ablation study}

\begin{figure}[t]
                \centering
                \includegraphics[width=\columnwidth]{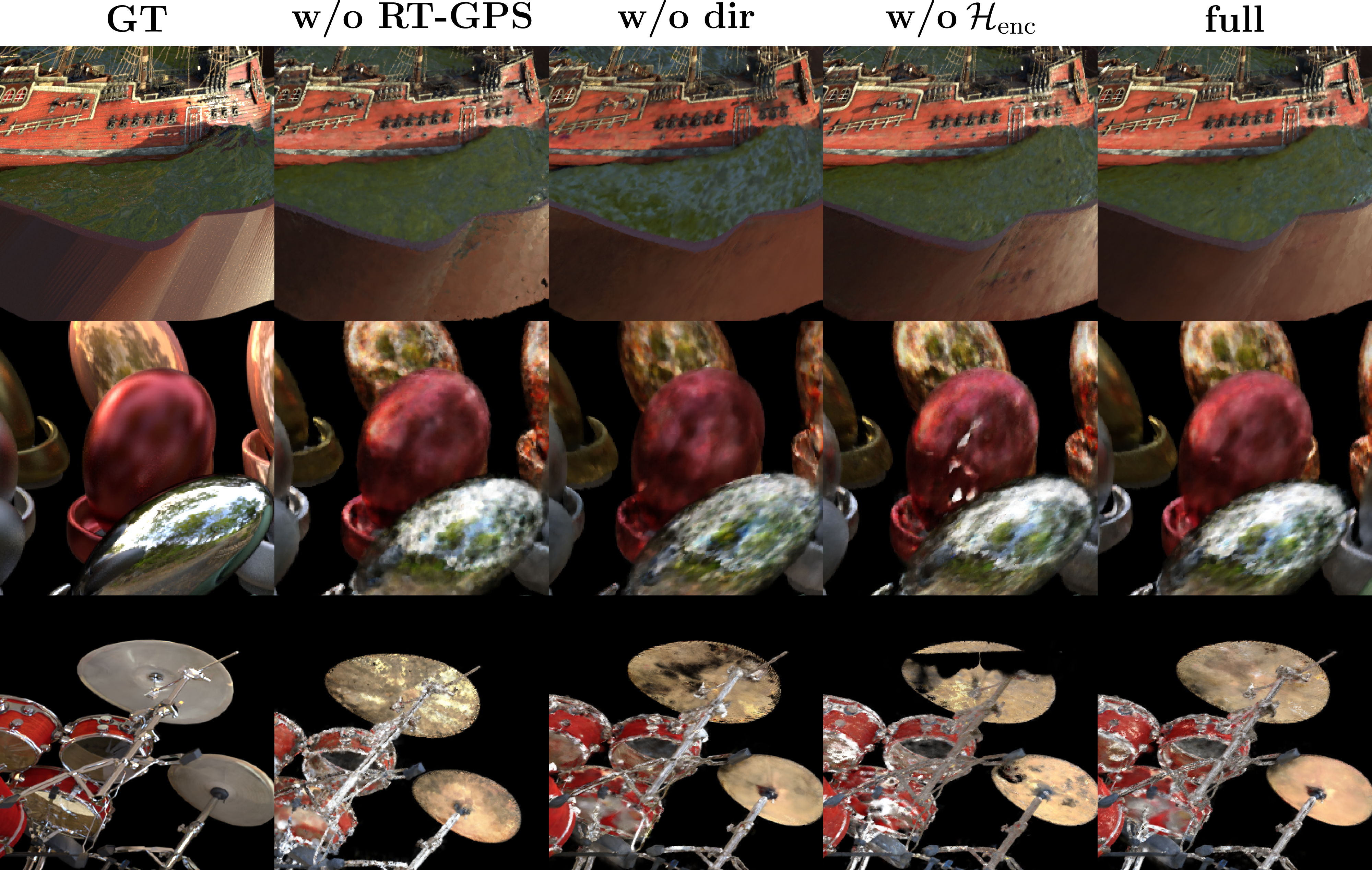}
                \caption{\textbf{Ablation study.} Qualitative comparison showing the effect of individual components on rendering quality.}
                \label{fig:ablation}
                \vspace{-0.2cm}
            \end{figure}

            We conduct an ablation study to assess the contribution of each major component of our method. We evaluate variants that (1) replace \ourknnshort{} with Euclidean KNN ($\text{w/o \ourknnshort{}}$), (2) remove hash-grid feature distillation and use learned per-Gaussian features ($\text{w/o}\:\mathcal{H}_{\text{enc}}$), and (3) disable view-direction restoration ($\text{w/o dir}$).
            For static reconstruction, all variants achieve comparable performance, with only minor PSNR differences as shown in Table~\ref{tab:ablation}. In contrast, edited-scene reconstruction is more sensitive to architectural choices. Removing view-direction restoration leads to the largest performance drop, as the model fails to recover correct view-dependent appearance after deformation. Using Euclidean KNN introduces artifacts similar to point-based baselines, while removing hash-grid feature distillation has a smaller quantitative impact.
            However, qualitative results in Figure~\ref{fig:ablation} reveal that omitting hash-grid distillation leads to visible artifacts, including holes and floating structures. Since these artifacts occur sparsely, their impact on PSNR remains limited, highlighting the importance of qualitative evaluation.

            \begin{table}[h!]
            \vspace{-0.5cm}
            {\small 
            \begin{center}
                \begin{tabular}{l@{\;}l@{\;}l@{\;}l@{\;}l@{\;}l@{\;}l@{\;}l@{\;}l@{\;}l@{}}
                     & \rot{Chair} & \rot{Drums} & \rot{Lego} & \rot{Mic} & \rot{Materials} & \rot{Ship} & \rot{Hotdog} & \rot{Ficus} \\ 
             \hline
            
                \multicolumn{9}{c}{Static Reconstruction} \\ 
             \hline
            w/o \ourknnshort{} & 34.24 & 25.83 & \bf 36.02 & 36.13 & 29.98 & 30.72 & 36.97 & 33.21 \\
            w/o $\mathcal{H}_\text{enc}$ & \bf 34.72 & 25.74 & 35.74 & 35.89 & 29.89 & 30.79 & 37.09 & \bf 33.98 \\
            full & \bf 34.72 & \bf 26.01 & 35.59 & \bf 36.54 & \bf 30.08 & \bf 31.10 & \bf 37.11 & 33.82 \\
            \hline
                \multicolumn{9}{c}{Editing} \\
            \hline
            w/o dir & 23.80 & 21.48 & 26.07 & 27.32 & 21.15 & 21.72 & 27.70 & 25.90 \\
            w/o \ourknnshort{} & 25.58 & 21.57 & 27.55 & 27.59 & 22.91 & 24.17 & 27.28 & 26.47 \\
            w/o $\mathcal{H}_\text{enc}$ & 25.98 & 21.83 & \bf 28.07 & 27.70 & 23.05 & \bf 24.51 & 28.03 & \bf 27.64 \\
            full & \bf 26.03 & \bf 22.08 & 28.04 & \bf 27.85 & \bf 23.14 & 24.43 & \bf 28.23 & 27.58 \\
            \hline

                \end{tabular}
                \end{center}
                }
                \caption{
                Ablation study of \our{} reporting PSNR for static reconstruction and edited scenes.
                }
                \label{tab:ablation}
                \vspace{-0.7cm}
            \end{table}

\section{Conclusions}

    We introduced \our{}, an affine-equivariant kernel space encoding for Neural Radiance Fields that enables stable, localized, and deformation-aware scene editing. By representing latent features with anisotropic Gaussian kernels and aggregating them using Mahalanobis-distance-based neighbourhoods, our method preserves local feature structure under affine transformations, addressing a key limitation of point- and grid-based NeRF encodings.
    To retain high reconstruction quality, we proposed a training-time hash-grid feature distillation mechanism that transfers multi-resolution grid features into a compact, grid-free kernel representation at inference. This allows \our{} to achieve reconstruction quality comparable to state-of-the-art NeRF models while enabling direct, intuitive editing without retraining. Across quantitative benchmarks and qualitative evaluations, our approach consistently outperforms prior editable NeRF methods, particularly after complex edits.
    Finally, we demonstrated that \our{} naturally supports physics-driven scene manipulation, enabling realistic rigid-body, soft-body, and cloth simulations when integrated with standard physics engines. These results suggest that kernel-based, transformation-aware latent encodings provide a promising foundation for physically interactive and editable neural scene representations.

\section{Acknowledgments} 
    The work of P. Spurek was supported by the National Centre of Science (Poland) Grant No. 2023/50/E/ST6/00068. The work of M. Zieli\'{n}ski was supported by the National Science Centre, Poland, under research project no UMO-2023/51/B/ST6/01646. Some of the computations presented in this work were carried out using the infrastructure of the Poznań Supercomputing and Networking Center (PCSS).

\bibliography{bibliography}

@inproceedings{wang2023seal3d,
  author = {Wang, Xiangyu and Zhu, Jingsen and Ye, Qi and Huo, Yuchi and Ran, Yunlong and Zhong, Zhihua and Chen, Jiming},
  booktitle = {ICCV},
  ee = {https://doi.org/10.1109/ICCV51070.2023.01621},
  isbn = {979-8-3503-0718-4},
  pages = {17637-17647},
  publisher = {IEEE},
  title = {Seal-3D: Interactive Pixel-Level Editing for Neural Radiance Fields.},
  year = 2023
}

@inproceedings{mildenhall2020nerf,
 title={{NeRF: Representing Scenes as Neural Radiance Fields for View Synthesis}},
 author={Ben Mildenhall and Pratul P. Srinivasan and Matthew Tancik and Jonathan T. Barron and Ravi Ramamoorthi and Ren Ng},
 year={2020},
 booktitle={ECCV},
}

@inproceedings{wang2023rip,
  title={Rip-nerf: Learning rotation-invariant point-based neural radiance field for fine-grained editing and compositing},
  author={Wang, Yuze and Wang, Junyi and Qu, Yansong and Qi, Yue},
  booktitle={Proceedings of the 2023 ACM international conference on multimedia retrieval},
  pages={125--134},
  year={2023}
}

@inproceedings{yuan2022nerf,
  title={Nerf-editing: geometry editing of neural radiance fields},
  author={Yuan, Yu-Jie and Sun, Yang-Tian and Lai, Yu-Kun and Ma, Yuewen and Jia, Rongfei and Gao, Lin},
  booktitle={Proceedings of the IEEE/CVF conference on computer vision and pattern recognition},
  pages={18353--18364},
  year={2022}
}

@inproceedings{chen2023neuraleditor,
  title={Neuraleditor: Editing neural radiance fields via manipulating point clouds},
  author={Chen, Jun-Kun and Lyu, Jipeng and Wang, Yu-Xiong},
  booktitle={Proceedings of the IEEE/CVF conference on computer vision and pattern recognition},
  pages={12439--12448},
  year={2023}
}

@inproceedings{yang2022neumesh,
  title={Neumesh: Learning disentangled neural mesh-based implicit field for geometry and texture editing},
  author={Yang, Bangbang and Bao, Chong and Zeng, Junyi and Bao, Hujun and Zhang, Yinda and Cui, Zhaopeng and Zhang, Guofeng},
  booktitle={European Conference on Computer Vision},
  pages={597--614},
  year={2022},
  organization={Springer}
}

@article{zhang2023papr,
  title={Papr: Proximity attention point rendering},
  author={Zhang, Yanshu and Peng, Shichong and Moazeni, Alireza and Li, Ke},
  journal={Advances in Neural Information Processing Systems},
  volume={36},
  pages={60307--60328},
  year={2023}
}

@article{borycki2024gasp,
  title={Gasp: Gaussian splatting for physic-based simulations},
  author={Borycki, Piotr and Smolak, Weronika and Waczy{\'n}ska, Joanna and Mazur, Marcin and Tadeja, S{\l}awomir and Spurek, Przemys{\l}aw},
  journal={arXiv preprint arXiv:2409.05819},
  year={2024}
}

@article{malarz2025gaussian,
  title={Gaussian splatting with nerf-based color and opacity},
  author={Malarz, Dawid and Smolak-Dy{\.z}ewska, Weronika and Tabor, Jacek and Tadeja, S{\l}awomir and Spurek, Przemys{\l}aw},
  journal={Computer Vision and Image Understanding},
  volume={251},
  pages={104273},
  year={2025},
  publisher={Elsevier}
}

@article{kerbl20233d,
  title={3D Gaussian Splatting for Real-Time Radiance Field Rendering},
  author={Kerbl, Bernhard and Kopanas, Georgios and Leimk{\"u}hler, Thomas and Drettakis, George},
  journal={ACM Transactions on Graphics},
  volume={42},
  number={4},
  year={2023}
}

@inproceedings{xie2024physgaussian,
  title={Physgaussian: Physics-integrated 3d gaussians for generative dynamics},
  author={Xie, Tianyi and Zong, Zeshun and Qiu, Yuxing and Li, Xuan and Feng, Yutao and Yang, Yin and Jiang, Chenfanfu},
  booktitle={Proceedings of the IEEE/CVF Conference on Computer Vision and Pattern Recognition},
  pages={4389--4398},
  year={2024}
}

@article{monnier2023differentiable,
  title={Differentiable blocks world: Qualitative 3d decomposition by rendering primitives},
  author={Monnier, Tom and Austin, Jake and Kanazawa, Angjoo and Efros, Alexei and Aubry, Mathieu},
  journal={Advances in Neural Information Processing Systems},
  volume={36},
  pages={5791--5807},
  year={2023}
}

@article{muller2022instant,
  title={Instant neural graphics primitives with a multiresolution hash encoding},
  author={M{\"u}ller, Thomas and Evans, Alex and Schied, Christoph and Keller, Alexander},
  journal={ACM Transactions on Graphics (ToG)},
  volume={41},
  number={4},
  pages={1--15},
  year={2022},
  publisher={ACM New York, NY, USA}
}

@inproceedings{govindarajan2024lagrangian,
  title={Lagrangian hashing for compressed neural field representations},
  author={Govindarajan, Shrisudhan and Sambugaro, Zeno and Shabanov, Akhmedkhan and Takikawa, Towaki and Rebain, Daniel and Sun, Weiwei and Conci, Nicola and Yi, Kwang Moo and Tagliasacchi, Andrea},
  booktitle={European Conference on Computer Vision},
  pages={183--199},
  year={2024},
  organization={Springer}
}

@inproceedings{tretschk2021non,
  title={Non-rigid neural radiance fields: Reconstruction and novel view synthesis of a dynamic scene from monocular video},
  author={Tretschk, Edgar and Tewari, Ayush and Golyanik, Vladislav and Zollh{\"o}fer, Michael and Lassner, Christoph and Theobalt, Christian},
  booktitle={Proceedings of the IEEE/CVF international conference on computer vision},
  pages={12959--12970},
  year={2021}
}

@article{park2021hypernerf,
  title={HyperNeRF: a higher-dimensional representation for topologically varying neural radiance fields},
  author={Park, Keunhong and Sinha, Utkarsh and Hedman, Peter and Barron, Jonathan T and Bouaziz, Sofien and Goldman, Dan B and Martin-Brualla, Ricardo and Seitz, Steven M},
  journal={ACM Transactions on Graphics (TOG)},
  volume={40},
  number={6},
  pages={1--12},
  year={2021},
  publisher={ACM New York, NY, USA}
}

@inproceedings{hofherr2023neural,
  title={Neural implicit representations for physical parameter inference from a single video},
  author={Hofherr, Florian and Koestler, Lukas and Bernard, Florian and Cremers, Daniel},
  booktitle={Proceedings of the IEEE/CVF Winter Conference on Applications of Computer Vision},
  pages={2093--2103},
  year={2023}
}

@article{jambon2023nerfshop,
  title={Nerfshop: Interactive editing of neural radiance fields},
  author={Jambon, Cl{\'e}ment and Kerbl, Bernhard and Kopanas, Georgios and Diolatzis, Stavros and Leimk{\"u}hler, Thomas and Drettakis, George},
  journal={Proceedings of the ACM on Computer Graphics and Interactive Techniques},
  volume={6},
  number={1},
  year={2023}
}

@article{yuan2023interactive,
  title={Interactive nerf geometry editing with shape priors},
  author={Yuan, Yu-Jie and Sun, Yang-Tian and Lai, Yu-Kun and Ma, Yuewen and Jia, Rongfei and Kobbelt, Leif and Gao, Lin},
  journal={IEEE Transactions on Pattern Analysis and Machine Intelligence},
  volume={45},
  number={12},
  pages={14821--14837},
  year={2023},
  publisher={IEEE}
}

@inproceedings{xu2022deforming,
  title={Deforming radiance fields with cages},
  author={Xu, Tianhan and Harada, Tatsuya},
  booktitle={European Conference on Computer Vision},
  pages={159--175},
  year={2022},
  organization={Springer}
}

@article{peng2022cagenerf,
  title={Cagenerf: Cage-based neural radiance field for generalized 3d deformation and animation},
  author={Peng, Yicong and Yan, Yichao and Liu, Shengqi and Cheng, Yuhao and Guan, Shanyan and Pan, Bowen and Zhai, Guangtao and Yang, Xiaokang},
  journal={Advances in Neural Information Processing Systems},
  volume={35},
  pages={31402--31415},
  year={2022}
}

@inproceedings{feng2024pie,
  title={Pie-nerf: Physics-based interactive elastodynamics with nerf},
  author={Feng, Yutao and Shang, Yintong and Li, Xuan and Shao, Tianjia and Jiang, Chenfanfu and Yang, Yin},
  booktitle={Proceedings of the IEEE/CVF Conference on Computer Vision and Pattern Recognition},
  pages={4450--4461},
  year={2024}
}

@inproceedings{garbin2024voltemorph,
  title={VolTeMorph: Real-time, Controllable and Generalizable Animation of Volumetric Representations},
  author={Garbin, Stephan J and Kowalski, Marek and Estellers, Virginia and Szymanowicz, Stanislaw and Rezaeifar, Shideh and Shen, Jingjing and Johnson, Matthew A and Valentin, Julien},
  booktitle={Computer Graphics Forum},
  volume={43},
  pages={e15117},
  year={2024},
  organization={Wiley Online Library}
}

@inproceedings{zheng2023editablenerf,
  title={Editablenerf: Editing topologically varying neural radiance fields by key points},
  author={Zheng, Chengwei and Lin, Wenbin and Xu, Feng},
  booktitle={Proceedings of the IEEE/CVF Conference on Computer Vision and Pattern Recognition},
  pages={8317--8327},
  year={2023}
}

@inproceedings{kania2022conerf,
  title={Conerf: Controllable neural radiance fields},
  author={Kania, Kacper and Yi, Kwang Moo and Kowalski, Marek and Trzci{\'n}ski, Tomasz and Tagliasacchi, Andrea},
  booktitle={Proceedings of the IEEE/CVF Conference on Computer Vision and Pattern Recognition},
  pages={18623--18632},
  year={2022}
}

@inproceedings{weng2022humannerf,
  title={Humannerf: Free-viewpoint rendering of moving people from monocular video},
  author={Weng, Chung-Yi and Curless, Brian and Srinivasan, Pratul P and Barron, Jonathan T and Kemelmacher-Shlizerman, Ira},
  booktitle={Proceedings of the IEEE/CVF conference on computer vision and pattern Recognition},
  pages={16210--16220},
  year={2022}
}

@article{qiao2022neuphysics,
  title={Neuphysics: Editable neural geometry and physics from monocular videos},
  author={Qiao, Yi-Ling and Gao, Alexander and Lin, Ming},
  journal={Advances in Neural Information Processing Systems},
  volume={35},
  pages={12841--12854},
  year={2022}
}

@inproceedings{du2021neural,
  title={Neural radiance flow for 4d view synthesis and video processing},
  author={Du, Yilun and Zhang, Yinan and Yu, Hong-Xing and Tenenbaum, Joshua B and Wu, Jiajun},
  booktitle={2021 IEEE/CVF International Conference on Computer Vision (ICCV)},
  pages={14304--14314},
  year={2021},
  organization={IEEE Computer Society}
}

@inproceedings{bao2023sine,
  title={Sine: Semantic-driven image-based nerf editing with prior-guided editing field},
  author={Bao, Chong and Zhang, Yinda and Yang, Bangbang and Fan, Tianxing and Yang, Zesong and Bao, Hujun and Zhang, Guofeng and Cui, Zhaopeng},
  booktitle={Proceedings of the IEEE/CVF Conference on Computer Vision and Pattern Recognition},
  pages={20919--20929},
  year={2023}
}

@article{dong2023vica,
  title={Vica-nerf: View-consistency-aware 3d editing of neural radiance fields},
  author={Dong, Jiahua and Wang, Yu-Xiong},
  journal={Advances in Neural Information Processing Systems},
  volume={36},
  pages={61466--61477},
  year={2023}
}

@article{kobayashi2022decomposing,
  title={Decomposing nerf for editing via feature field distillation},
  author={Kobayashi, Sosuke and Matsumoto, Eiichi and Sitzmann, Vincent},
  journal={Advances in neural information processing systems},
  volume={35},
  pages={23311--23330},
  year={2022}
}

@inproceedings{jiang2022nerffaceediting,
  title={Nerffaceediting: Disentangled face editing in neural radiance fields},
  author={Jiang, Kaiwen and Chen, Shu-Yu and Liu, Feng-Lin and Fu, Hongbo and Gao, Lin},
  booktitle={SIGGRAPH Asia 2022 Conference Papers},
  pages={1--9},
  year={2022}
}

@inproceedings{sun2022fenerf,
  title={Fenerf: Face editing in neural radiance fields},
  author={Sun, Jingxiang and Wang, Xuan and Zhang, Yong and Li, Xiaoyu and Zhang, Qi and Liu, Yebin and Wang, Jue},
  booktitle={Proceedings of the IEEE/CVF conference on computer vision and pattern recognition},
  pages={7672--7682},
  year={2022}
}

@inproceedings{lazova2023control,
  title={Control-nerf: Editable feature volumes for scene rendering and manipulation},
  author={Lazova, Verica and Guzov, Vladimir and Olszewski, Kyle and Tulyakov, Sergey and Pons-Moll, Gerard},
  booktitle={Proceedings of the IEEE/CVF Winter Conference on Applications of Computer Vision},
  pages={4340--4350},
  year={2023}
}

@inproceedings{hwang2023faceclipnerf,
  title={Faceclipnerf: Text-driven 3d face manipulation using deformable neural radiance fields},
  author={Hwang, Sungwon and Hyung, Junha and Kim, Daejin and Kim, Min-Jung and Choo, Jaegul},
  booktitle={Proceedings of the IEEE/CVF International Conference on Computer Vision},
  pages={3469--3479},
  year={2023}
}

@inproceedings{weder2023removing,
  title={Removing objects from neural radiance fields},
  author={Weder, Silvan and Garcia-Hernando, Guillermo and Monszpart, Aron and Pollefeys, Marc and Brostow, Gabriel J and Firman, Michael and Vicente, Sara},
  booktitle={Proceedings of the IEEE/CVF Conference on Computer Vision and Pattern Recognition},
  pages={16528--16538},
  year={2023}
}

@inproceedings{song2023blending,
  title={Blending-nerf: Text-driven localized editing in neural radiance fields},
  author={Song, Hyeonseop and Choi, Seokhun and Do, Hoseok and Lee, Chul and Kim, Taehyeong},
  booktitle={Proceedings of the IEEE/CVF international conference on computer vision},
  pages={14383--14393},
  year={2023}
}

@inproceedings{srinivasan2021nerv,
  title={Nerv: Neural reflectance and visibility fields for relighting and view synthesis},
  author={Srinivasan, Pratul P and Deng, Boyang and Zhang, Xiuming and Tancik, Matthew and Mildenhall, Ben and Barron, Jonathan T},
  booktitle={Proceedings of the IEEE/CVF conference on computer vision and pattern recognition},
  pages={7495--7504},
  year={2021}
}

@inproceedings{rudnev2022nerf,
  title={Nerf for outdoor scene relighting},
  author={Rudnev, Viktor and Elgharib, Mohamed and Smith, William and Liu, Lingjie and Golyanik, Vladislav and Theobalt, Christian},
  booktitle={European Conference on Computer Vision},
  pages={615--631},
  year={2022},
  organization={Springer}
}

@inproceedings{liu2021editing,
  title={Editing conditional radiance fields},
  author={Liu, Steven and Zhang, Xiuming and Zhang, Zhoutong and Zhang, Richard and Zhu, Jun-Yan and Russell, Bryan},
  booktitle={Proceedings of the IEEE/CVF international conference on computer vision},
  pages={5773--5783},
  year={2021}
}

@inproceedings{gong2023recolornerf,
  title={Recolornerf: Layer decomposed radiance fields for efficient color editing of 3d scenes},
  author={Gong, Bingchen and Wang, Yuehao and Han, Xiaoguang and Dou, Qi},
  booktitle={Proceedings of the 31st ACM International Conference on Multimedia},
  pages={8004--8015},
  year={2023}
}

@inproceedings{wang2022clip,
  title={Clip-nerf: Text-and-image driven manipulation of neural radiance fields},
  author={Wang, Can and Chai, Menglei and He, Mingming and Chen, Dongdong and Liao, Jing},
  booktitle={Proceedings of the IEEE/CVF conference on computer vision and pattern recognition},
  pages={3835--3844},
  year={2022}
}

@inproceedings{haque2023instruct,
  title={Instruct-nerf2nerf: Editing 3d scenes with instructions},
  author={Haque, Ayaan and Tancik, Matthew and Efros, Alexei A and Holynski, Aleksander and Kanazawa, Angjoo},
  booktitle={Proceedings of the IEEE/CVF international conference on computer vision},
  pages={19740--19750},
  year={2023}
}

@inproceedings{mikaeili2023sked,
  title={Sked: Sketch-guided text-based 3d editing},
  author={Mikaeili, Aryan and Perel, Or and Safaee, Mehdi and Cohen-Or, Daniel and Mahdavi-Amiri, Ali},
  booktitle={Proceedings of the IEEE/CVF International Conference on Computer Vision},
  pages={14607--14619},
  year={2023}
}

@inproceedings{cao2023hexplane,
  title={Hexplane: A fast representation for dynamic scenes},
  author={Cao, Ang and Johnson, Justin},
  booktitle={Proceedings of the IEEE/CVF Conference on Computer Vision and Pattern Recognition},
  pages={130--141},
  year={2023}
}

@inproceedings{guo2023forward,
  title={Forward flow for novel view synthesis of dynamic scenes},
  author={Guo, Xiang and Sun, Jiadai and Dai, Yuchao and Chen, Guanying and Ye, Xiaoqing and Tan, Xiao and Ding, Errui and Zhang, Yumeng and Wang, Jingdong},
  booktitle={Proceedings of the IEEE/CVF International Conference on Computer Vision},
  pages={16022--16033},
  year={2023}
}

@inproceedings{gao2021dynamic,
  title={Dynamic view synthesis from dynamic monocular video},
  author={Gao, Chen and Saraf, Ayush and Kopf, Johannes and Huang, Jia-Bin},
  booktitle={Proceedings of the IEEE/CVF International Conference on Computer Vision},
  pages={5712--5721},
  year={2021}
}

@inproceedings{park2021nerfies,
  title={Nerfies: Deformable neural radiance fields},
  author={Park, Keunhong and Sinha, Utkarsh and Barron, Jonathan T and Bouaziz, Sofien and Goldman, Dan B and Seitz, Steven M and Martin-Brualla, Ricardo},
  booktitle={Proceedings of the IEEE/CVF international conference on computer vision},
  pages={5865--5874},
  year={2021}
}

@inproceedings{xu2022point,
  title={Point-nerf: Point-based neural radiance fields},
  author={Xu, Qiangeng and Xu, Zexiang and Philip, Julien and Bi, Sai and Shu, Zhixin and Sunkavalli, Kalyan and Neumann, Ulrich},
  booktitle={CVPR},
  pages={5438--5448},
  year={2022}
}

@article{Nagarajan2023RTKNNS,
  author    = {Vani Nagarajan and
               Durga Mandarapu and
               Milind Kulkarni},
  title     = {{RT-kNNS Unbound: Using RT Cores to Accelerate Unrestricted Neighbor Search}},
  journal   = {CoRR},
  volume    = {abs/2305.18356},
  year      = {2023},
  url       = {https://arxiv.org/abs/2305.18356},
  eprint    = {2305.18356},
  archivePrefix = {arXiv},
  primaryClass = {cs.LG},
  note      = {Accepted at the International Conference on Supercomputing 2023 (ICS'23)}
}

@article{barron2022mipnerf360,
    title={Mip-NeRF 360: Unbounded Anti-Aliased Neural Radiance Fields},
    author={Jonathan T. Barron and Ben Mildenhall and 
            Dor Verbin and Pratul P. Srinivasan and Peter Hedman},
    journal={CVPR},
    year={2022}
}

@Manual{blender,
   title = {Blender - a 3D modelling and rendering package},
   author = {Blender Online Community},
   organization = {Blender Foundation},
   address = {Stichting Blender Foundation, Amsterdam},
   year = {2018},
   url = {http://www.blender.org},
}

@Manual{Genesis,
          author = {Genesis Authors},
          title = {Genesis: A Universal and Generative Physics Engine for Robotics and Beyond},
          month = {December},
          year = {2024},
          url = {https://github.com/Genesis-Embodied-AI/Genesis}
        }

@article{huang2024blenderalchemy,
  title={BlenderAlchemy: Editing 3D Graphics with Vision-Language Models},
  author={Huang, Ian and Yang, Guandao and Guibas, Leonidas},
  journal={arXiv preprint arXiv:2404.17672},
  year={2024}
}

@inproceedings{weber2024nerfiller,
  author = {Weber, Ethan and Holynski, Aleksander and Jampani, Varun and Saxena, Saurabh and Snavely, Noah and Kar, Abhishek and Kanazawa, Angjoo},
  booktitle = {CVPR},
  isbn = {979-8-3503-5300-6},
  pages = {20731-20741},
  publisher = {IEEE},
  title = {NeRFiller: Completing Scenes via Generative 3D Inpainting.},
  year = 2024
}
\bibliographystyle{icml2026}

\newpage
\appendix
\onecolumn

    \section{Appendix}

    This appendix provides additional insights and supporting material for our method. We give a formal justification of the $k$-nearest neighbour approximation used in \ourknn, showing that distant Gaussians can be safely ignored with bounded error. We also provide extended qualitative and quantitative results to showcase the quality of our approach across various scenes.
    
    \section{Theoretical Motivation for \ourknn{} Approximation}
        \label{adx:theoretical}

        To justify the motivation behind our \textit{\ourknn}, let's first recall the formula for the interpolated feature vector $\mathbf{v}(\mathbf{x})$. To begin, let's note that for the $ w_i(\mathbf{x})$ appearing in the formula we have:
            $$
            w_i(\mathbf{x}) = 
                \begin{cases}
                \exp \left( -\frac{1}{2} d_M^2 \left( \mathbf{x}, \mathcal{N} \left( \boldsymbol{\mu}_i, \boldsymbol{\Sigma}_i \right) \right) \right), & \text{if } i \in N \\
                0, & \text{otherwise},
                \end{cases}
            $$
            where $d_M \left( \mathbf{x}, \mathcal{N} \left( \boldsymbol{\mu}_i, \boldsymbol{\Sigma}_i \right) \right)$ is the Mahalanobis distance of the point $\mathbf{x}$ from the normal distribution $\mathcal{N} \left( \boldsymbol{\mu}_i, \boldsymbol{\Sigma}_i \right)$. Let's fix $\mathbf{x} \in \mathbb{R}^3$ and $\varepsilon > 0$. Let's consider the subset $M \subseteq N$, such that for each $i \in M$ we have:
            $$
            d_M \left( \mathbf{x}, \mathcal{N} \left( \boldsymbol{\mu}_i, \boldsymbol{\Sigma}_i \right) \right) > \sqrt{-2 \ln \left( \frac{ \varepsilon}{\sum \limits_{i \in M} \left \lvert \mathbf{v(\mathbf{x})}_i \right \rvert} \right) }
            $$
            Then:
            \begin{flalign*}
            & \left \lvert \sum \limits_{i \in M} w_i(\mathbf{x}) \cdot v(\mathbf{x})_i \right \rvert =&\\
            &= \left \lvert \sum \limits_{i \in M} e^{-\frac{1}{2} d_M^2 \left( \mathbf{x}, \mathcal{N} \left( \boldsymbol{\mu}_i, \boldsymbol{\Sigma}_i \right) \right)} \cdot v(\mathbf{x})_i \right \rvert \le&\\
            &\le \sum \limits_{i \in M} \left \lvert e^{-\frac{1}{2} d_M^2 \left( \mathbf{x}, \mathcal{N} \left( \boldsymbol{\mu}_i, \boldsymbol{\Sigma}_i \right) \right)} \right \rvert \cdot \left \lvert v(\mathbf{x})_i \right \rvert =&\\
            &= e^{-\frac{1}{2} d_M^2 \left( \mathbf{x}, \mathcal{N} \left( \boldsymbol{\mu}_i, \boldsymbol{\Sigma}_i \right) \right)} \cdot \sum \limits_{i \in M} \left \lvert \mathbf{v(\mathbf{x})}_i \right \rvert <&\\
            &< \frac{\varepsilon}{\sum \limits_{i \in M} \left \lvert \mathbf{v(\mathbf{x})}_i \right \rvert} \cdot \sum \limits_{i \in M} \left \lvert \mathbf{v(\mathbf{x})}_i \right \rvert = \varepsilon
            \end{flalign*}
            Thus:
            \begin{flalign*}
            & \left \lvert \sum \limits_{i \in N} w_i(\mathbf{x}) \cdot v(\mathbf{x})_i - \sum \limits_{i \in {N \setminus M}} w_i(\mathbf{x}) \cdot v(\mathbf{x})_i \right \rvert =&\\
            &= \left \lvert \sum \limits_{i \in M} w_i(\mathbf{x}) \cdot v(\mathbf{x})_i \right \rvert < \varepsilon
            \end{flalign*}
            from which we conclude that removing the nearest neighbors from the set $M$ from the formula for $\mathbf{v}\left(\mathcal{G}_{\our}\right)$ can alter the interpolated feature vector coordinate by no more than $\varepsilon$.

        \section{Extended results}

        In this section, we extend the results presented in Table 1 of the main paper by additionally reporting SSIM and LPIPS metrics for both synthetic and real-world datasets.

        \begin{table*}[ht]
        {\small 
        \begin{center}
            \begin{tabular}{l@{\quad}c@{\quad}c@{\quad}c@{\quad}c@{\quad}c@{\quad}c@{\quad}c@{\quad}c@{\quad}c@{\quad}}

        \multicolumn{9}{c}{PSNR $\uparrow$} \\ 
                 & Chair & Drums & Lego & Mic & Materials & Ship & Hotdog & Ficus \\ 
         \hline
        
        \multicolumn{9}{c}{Static} \\ 
        \hline
        
        INGP & 31.97 & 22.67 & 33.44 & 31.38 & 22.66 & 28.83 & 34.04 & 29.47 \\
        LagHash & \bf 35.66 & \bf 25.68 & \bf 35.49 & \bf 36.71 & \bf 29.60 & \bf 30.88 & \bf 37.30 & \bf 33.83 \\
        
        \hline
        \multicolumn{9}{c}{Editable} \\ 
        \hline
        GaMeS & \bf 35.73 & 26.15  & 35.57  & 35.67  & 29.89  & 30.78  & \bf 37.58  & 34.83 \\
        RIP-NeRF & 34.84 & 24.89 & 33.41 & 34.19 &28.31 & 30.65 & 35.96 & 32.23 \\
        Point-NeRF & 35.40 & 26.06 & 35.04 & 35.95 & 29.61 & 30.97 & 37.30 & \bf 36.13 \\
        Neuraleditor & 34.94 & \bf 26.19 & 34.28 & 36.09 & \bf 30.38 & 29.99 & 36.70 & 33.64 \\
        \our{} & 34.72 & 26.01 & \bf 35.59 & \bf 36.54 & 30.08 & \bf 31.10 & 37.11 & 33.82 \\ 
        
        \multicolumn{9}{c}{} \\ 
        \multicolumn{9}{c}{} \\ 
        
        \multicolumn{9}{c}{SSIM $\uparrow$} \\ 
        \hline
        \multicolumn{9}{c}{Static} \\
        \hline
        
        INGP & 0.976 & 0.900 & 0.974 & 0.975 & 0.889 & 0.860 & 0.976 & 0.962 \\
        LagHash & \bf 0.984 & \bf 0.934 & \bf 0.978 & \bf 0.991 & \bf 0.947 & \bf 0.892 & \bf 0.981 & \bf 0.981 \\
        
        \hline
            \multicolumn{9}{c}{Editable} \\ 
        \hline
        
        GaMeS & 0.987 & 0.953 & 0.982 & 0.992 & 0.952 & 0.904 & 0.985 & 0.986 \\
        RIP-NeRF & 0.980 & 0.929 & 0.977 & 0.962 & 0.943 & 0.916 & 0.963 & 0.979 \\
        Point-NeRF & \bf 0.991 & \bf 0.954 & \bf 0.988 & \bf 0.994 & \bf 0.971 & \bf 0.942 & \bf 0.991 & \bf 0.993 \\
        Neuraleditor & 0.980 & 0.928 & 0.974 & 0.985 & 0.960 & 0.876 & 0.969 & 0.970 \\
        \our{} & 0.983 & 0.939 & 0.978 & 0.990 & 0.951 & 0.898 & 0.981 & 0.977 \\ 
        \hline

        \multicolumn{9}{c}{} \\ 
        \multicolumn{9}{c}{} \\ 
        
        \multicolumn{9}{c}{LPIPS $\downarrow$} \\ 
        \hline
        \multicolumn{9}{c}{Static} \\ 
        \hline
        
        INGP & \bf 0.017 & 0.094 & \bf 0.013 & 0.027 & 0.100 & \bf 0.119 & \bf 0.021 & \bf 0.030 \\
        LagHash & 0.024 & \bf 0.083 & 0.027 & \bf 0.015 & \bf 0.070 & 0.139 & 0.036 & 0.049 \\
        
        \hline
            \multicolumn{9}{c}{Editable} \\ 
        \hline
        
        GaMeS & \bf 0.009 & \bf 0.038 & 0.014 & \bf 0.005 & 0.042 & 0.090 & 0.017 & 0.012 \\
        RIP-NeRF & - & - & - & - & - & - & - & - \\
        Point-NeRF & 0.010 & 0.055 & \bf 0.011 & 0.007 & 0.041 & \bf 0.070 & \bf 0.016 & \bf 0.009 \\
        Neuraleditor & 0.019 & 0.061 & 0.019 & 0.016 & \bf 0.031 & 0.075 & 0.033 & 0.033 \\
        \our{} & 0.011 & 0.052 & \bf 0.011 & 0.008 & 0.036 & 0.095 & \bf 0.016 & 0.015  \\ 
        \hline

            \end{tabular}
            \end{center}
            }
            \caption{Quantitative comparisons (PSNR, SSIM, LPIPS) on a NeRF-Synthetic dataset showing that \our{} gives comparable results with other models.}
            \label{tab:nerf_synthetic_extended}
        \end{table*}

    \section{High resolution qualitative results}
    \label{adx:enlarged}

    In this section we present enlarged qualitative comparison images of our method.

    \begin{figure}[t]
        \centering
        \includegraphics[width=0.9\columnwidth]{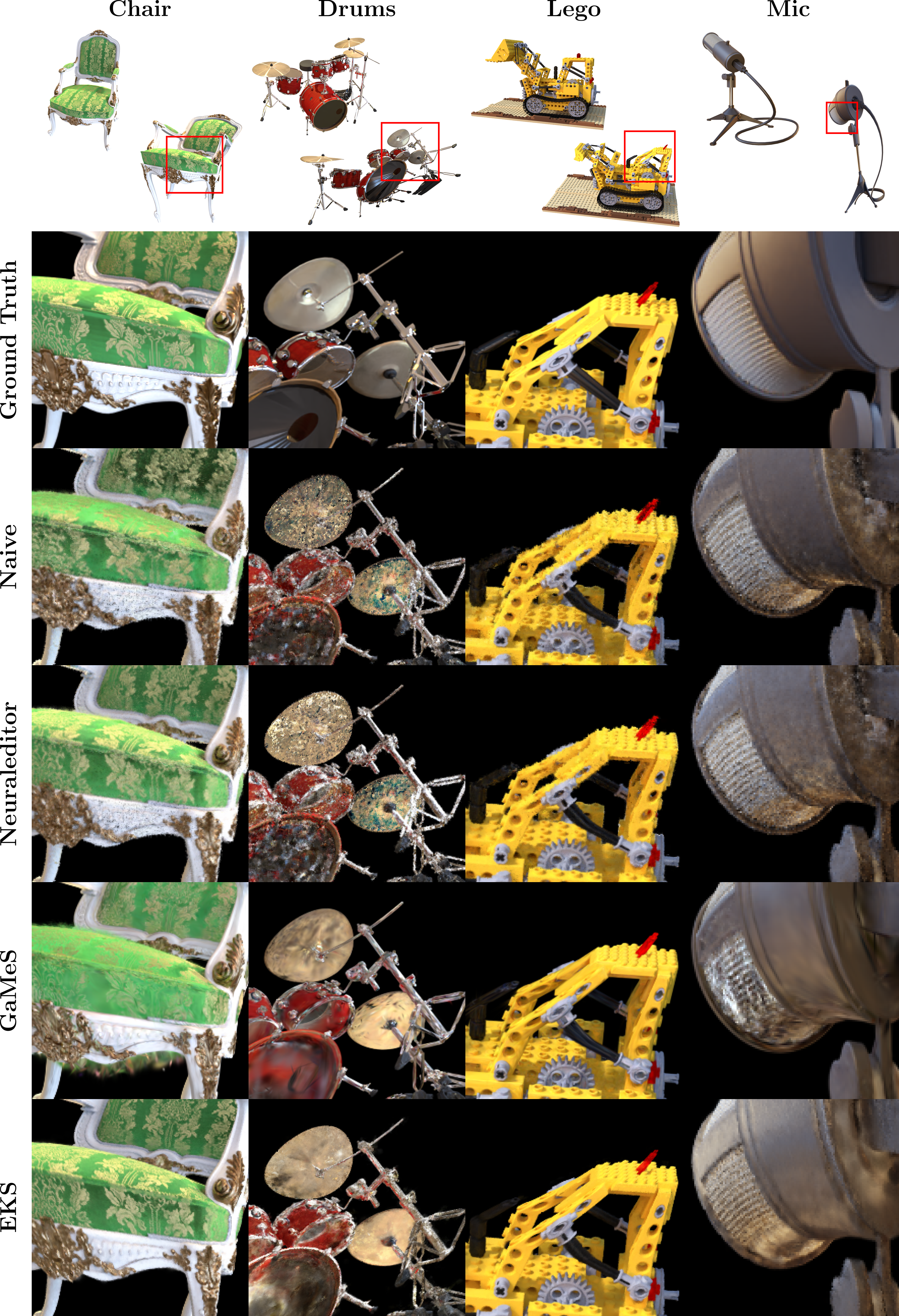}
        \caption{\textbf{Qualitative comparison.} Modified objects are in the top row. Each row compares reconstruction quality across different methods.}
    \end{figure}

    \begin{figure}[t]
        \centering
        \includegraphics[width=0.9\columnwidth]{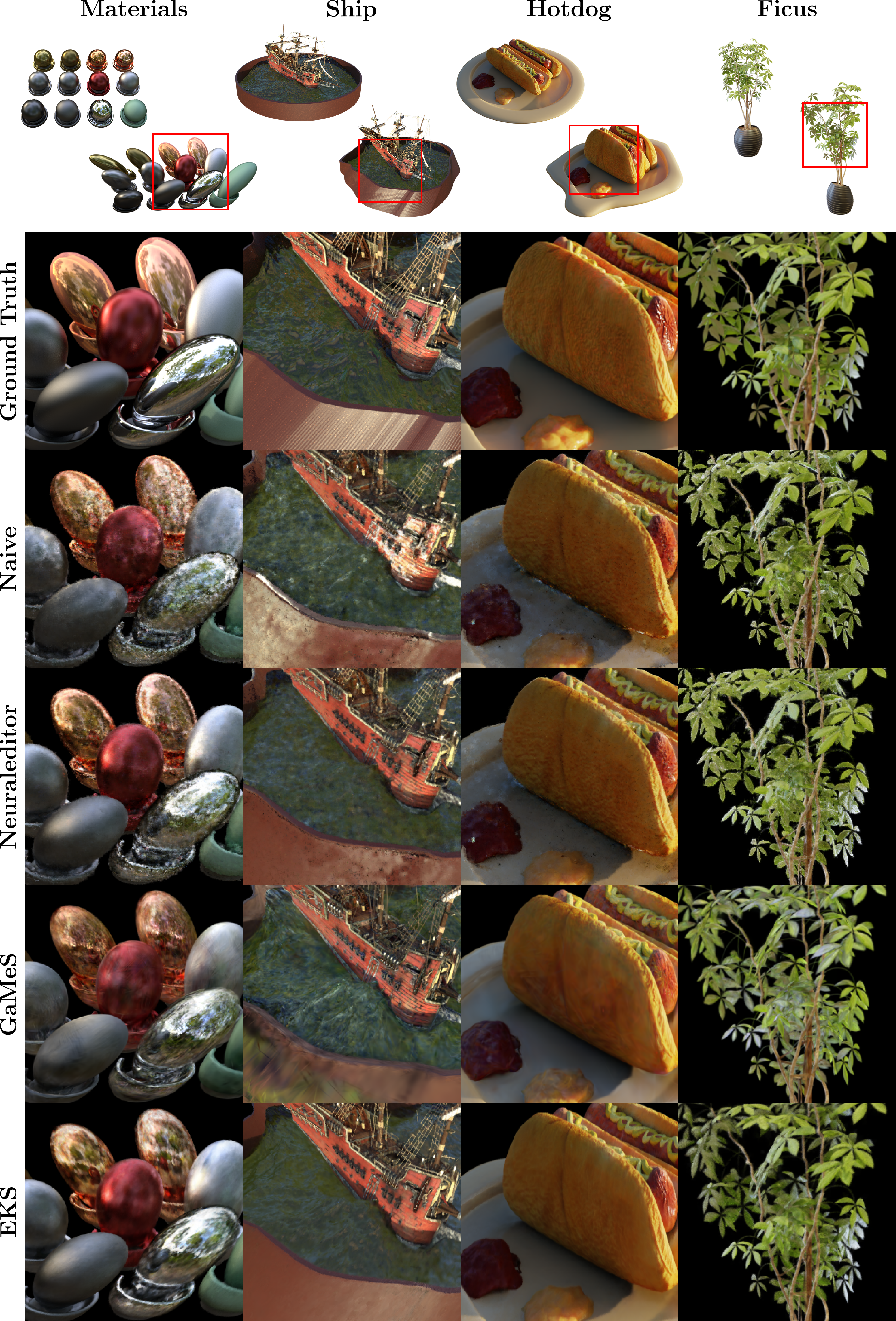}
        \caption{\textbf{Qualitative comparison.} Modified objects are in the top row. Each row compares reconstruction quality across different methods.}
    \end{figure}


\end{document}